\documentclass[10pt,journal,nodraftcls,compsoc]{IEEEtran}
\ifCLASSOPTIONcompsoc
  \usepackage[nocompress]{cite}

\usepackage{times}
\usepackage{epsfig}
\usepackage{graphicx}
\usepackage{amsmath}
\usepackage{amssymb}
\usepackage{verbatim}
\usepackage{booktabs} 
\usepackage{adjustbox}
\usepackage{colortbl} 
\usepackage{xcolor} 
\usepackage{xfrac}
\usepackage[]{algorithm2e}
\usepackage{subcaption}
\usepackage{graphicx,multirow}
\usepackage{array}
\usepackage{colortbl}
\usepackage{soul}
\usepackage{subcaption}
\usepackage{soul}

\usepackage{booktabs,subcaption,amsfonts,dcolumn}
\newcolumntype{d}[1]{D..{#1}}

\ifCLASSOPTIONcompsoc
  \usepackage[nocompress]{cite}
\else
  \usepackage{cite}
\fi
\ifCLASSINFOpdf
\else
\fi
\begin{document}
%
\title{Quality-Aware Multimodal Biometric Recognition}
%
%
%
%

\author{Sobhan Soleymani, Ali Dabouei, Fariborz Taherkhani, Seyed Mehdi Iranmanesh,\\Jeremy Dawson and Nasser M. Nasrabadi,~\IEEEmembership{Fellow,~IEEE}\\Lane Department of Computer Science and Electrical Engineering\\
West Virginia University}

\markboth{IEEE TRANSACTIONS ON BIOMETRICS, BEHAVIOR, AND IDENTITY SCIENCE}%
{Shell \MakeLowercase{\textit{et al.}}: Bare Demo of IEEEtran.cls for IEEE Journals}

\IEEEtitleabstractindextext{%
\begin{abstract}
We present a quality-aware multimodal recognition framework that combines representations from multiple biometric traits with varying quality and number of samples to achieve increased recognition accuracy by extracting complimentary identification information based on the quality of the samples. We develop a quality-aware framework for fusing representations of input modalities by weighting their importance using quality scores estimated in a weakly-supervised fashion. This framework utilizes two fusion blocks, each represented by a set of quality-aware and aggregation networks. In addition to architecture modifications, we propose two task-specific loss functions: multimodal separability loss and multimodal compactness loss. The first loss assures that the representations of modalities for a class have comparable magnitudes to provide a better quality estimation, while the multimodal representations of different classes are distributed to achieve maximum discrimination in the embedding space. The second loss, which is considered to regularize the network weights, improves the generalization performance by regularizing the framework. We evaluate the performance by considering three multimodal datasets consisting of face, iris, and fingerprint modalities. The efficacy of the framework is demonstrated through comparison with the state-of-the-art algorithms. In particular, our framework outperforms the rank- and score-level fusion of modalities of BIOMDATA~\cite{crihalmeanu2007protocol} by more than $30\%$ for true acceptance rate at false acceptance rate of $10^{-4}$.
\end{abstract}

}

\maketitle

\IEEEdisplaynontitleabstractindextext

%
\IEEEpeerreviewmaketitle

\IEEEraisesectionheading{\section{Introduction}\label{sec:introduction}}
\IEEEPARstart{B}iometrics research explores the possibility of automatically recognizing individuals based on their unique physical or behavioral traits such as face, fingerprint, voice, iris, or handwriting, which are referred to as biometric modalities. The uniqueness of the biometric features extracted from these traits, have allowed unimodal biometric systems to be widely used for identification and verification applications in a wide variety of scenarios~\cite{haghighat2016discriminant,bahrampour2016multimodal,shekhar2014joint,sundararajan2018deep,shekhar2014joint}. Beyond unimodal systems, a major merit of multimodal biometric recognition is its robustness to noisy data, non-universality, and category-based variations. Indeed, fusing multiple instances of biometric information lowers recognition error rates for low-quality and unreliable biometric samples, such as latent fingerprints, face images captured at a distance, and low-resolution iris images~\cite{singh2019comprehensive,jaafar2013review,toli2014survey}. Hence, we argue that the multimodal framework should be able to leverage the quality information of the input samples to incorporate all identification information within these samples while discarding the distorted information in low-quality samples that may negatively affect the identification.
\begin{figure*}[t]
\begin{center}
\includegraphics[width=.95\linewidth]{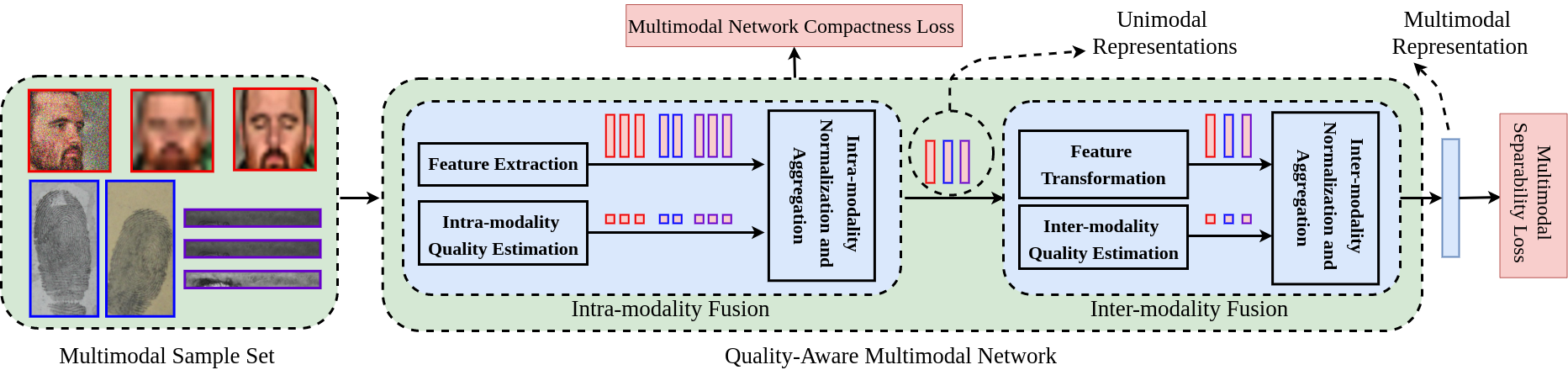}
\end{center}
\caption{A multimodal biometric sample set consists of samples from different modalities and varying quality. The quality-aware multimodal network, which consists of two fusion blocks, represents the sample set as a multimodal representation. The intra-modality and inter-modality qualities are estimated in a weakly-supervised fashion by minimizing the multimodal separability training loss, while the multimodal network compactness loss regularizes the network to provide better generalization.}  
\label{fig:MDN_0}
\end{figure*}

The most commonly deployed feature fusion methods for multimodal frameworks presented in the literature are feature concatenation~\cite{nagar2012multibiometric,goswami2016group}, bilinear multiplication~\cite{lin2015bilinear,chowdhury2016one}, and compact bilinear pooling~\cite{gao2016compact,delbrouck2017multimodal,fukui2016multimodal}. However, these methods treat all samples equally, and do not take their reliability and usefulness into account. A multimodal recognition algorithm requires selecting the discriminative and informative features from each sample, as well as exploring the dependencies between features extracted from different samples in a multimodal sample set. This framework should also determine the priority of features according to their usefulness and reliability for the recognition task. In addition, the information provided by different samples in a multimodal biometric sample set may or may not be independent. For instance, consider the case where the multimodal biometric recognition framework has access to videos captured by a surveillance camera and fingerprints collected using a sensor. In this situation, face images captured with pose variations are correlated, while the face images and fingerprint samples are independent. Therefore, compared to other recognition frameworks, information fusion for multimodal biometric systems has remained a challenging task.

A biometric sample is of good quality if it is suitable for automated matching. This quality can be quantified as a measure of how properly the biometric sample can be processed within the matching algorithm, including feature extraction and accurate recognition with a high confidence score~\cite{singh2019comprehensive}. For a multimodal sample set consisting of different modalities and a varying number of samples from each modality, the recognition framework should investigate both intra-modality and inter-modality information fusion. The intra-modality and inter-modality usefulness can be interpreted as the intra-modality quality of the samples and the inter-modality quality of different modalities, respectively.

Recent multi-sample recognition frameworks~\cite{liu2017quality,yang2017neural,zhao2017unconstrained,si2018dual} aim to solve this problem when all samples in the multimodal sample set are of the same modality. However, multimodal multi-sample recognition requires the consideration of independent samples in the set, where samples represent different modalities. Our proposed framework seeks to formalize a learning framework that automatically identifies the usefulness of the samples in a multimodal sample set through the loss defined by the underlying recognition task, where this usefulness is due to the intra-modality and inter-modality quality of the samples. This quality-aware framework aims to improve the representation of a multimodal sample set in the embedding space by estimating the quality of each of its samples in a weakly-supervised fashion. 

As presented in Fig.~\ref{fig:MDN_0}, our framework employs two weakly-supervised quality-aware fusion blocks. This framework represents each sample in the multimodal sample set with a feature vector and an intra-modality quality score. The feature vectors and quality scores associated with samples from each modality are utilized to construct a unimodal feature representation for each modality, while the inter-modality quality scores are utilized for credit assignment among the unimodal feature vectors in the fusion of the modalities. No quality scores are explicitly provided to the framework, and quality estimation allows different features from different samples to dynamically come to the forefront as needed by re-weighting the features when constructing the multimodal embedding space~\cite{xu2015show}. 

Our proposed multimodal recognition model includes two quality-aware fusion blocks for adaptive feature-level re-weighting~\cite{liu2017quality,yang2017neural}. We jointly train these quality-aware fusion blocks through minimizing {\it multimodal separability loss} and {\it  multimodal network compactness losses}. The first loss imposes an equi-distributed multimodal embedding in which the inter-class distance of multimodal representations is maximized and the intra-class variance is minimized. In addition, this loss function considers other constraints on the unimodal embeddings, linking them to the multimodal embedding. Our trained multimodal network is utilized during the test phase to incorporate the usefulness of each sample for the recognition task. Therefore, we propose multimodal network compactness loss to improves the generalization capability of the network by minimizing the hyperspherical energy for the layers of the network.

In the experimental setup, we focus on three multimodal recognition scenarios. In the first scenario, which can represent a traditional biometric framework, each modality in the multimodal sample set consists of a single sample. Here, the first quality-aware fusion block acts as a feature extraction block while the second one aims to satisfy the recognition task by learning the inter-modality quality of each modality. This scenario enables us to analyze the performance of the second fusion block. The second recognition scenario characterizes a multi-biometric capture system used for criminal booking. Due to variations in sensor type, training level of the booking officer, and overall human error, the biometric samples collected during booking and subsequently entered into an Electronic Biometric Transmission Specification (EBTS) can vary wildly in quality.  In this scenario, for each subject a set of low-quality face images and varying number of latent fingerprints are considered for the identification. This scenario mainly provides the possibility of studying the performance of the intra-modality fusion block. The third scenario, which can model an access control security system, is focused on representing a set of samples per modality with a single embedding space representation while the first fusion block considers the quality of each sample. Then, the second block aggregates the representations corresponding to different modalities through their inter-modality quality in the recognition task. This setup allows a more comprehensive evaluation of the joint performance of the two quality-aware fusion blocks.

The contributions of this paper in the field of multimodal biometrics are as follows:
\begin{itemize}
\item We propose a quality-aware fusion framework for multimodal biometrics applications which is optimized through learning in a weakly-supervised fashion without direct supervision of the quality of the samples or modalities.
\item We formalize the multi-sample multimodal recognition problem by learning two consecutive embeddings dedicated to extract discriminative features by exploiting the intra- and inter-modality information.  
\item An end-to-end training framework is proposed consisting of two novel loss functions for training the quality-aware fusion blocks.
\item Three {specific multi-sample multimodal person recognition scenarios are designed to carefully evaluate the performance of the proposed framework.} These scenarios consider two chimeric multimodal and one real-world multimodal datasets. 
\end{itemize}

\section{Background}
\subsection{Feature extraction and fusion}
Convolutional neural networks (CNNs) are efficient tools that can be employed to extract and represent discriminative features from raw data. Compared to hand-crafted features, the use of CNNs as domain feature extractors has been demonstrated to be more promising when facing different biometric modalities such as face~\cite{soleymani2018generalized}, iris~\cite{gangwar2016deepirisnet}, and fingerprint~\cite{nogueira2016fingerprint}. One of the major challenges in multimodal fusion is managing the large dimensionality of the fused feature representations, which highlights the importance of the fusion algorithm. In comparison to score-~\cite{mehrotra2016incremental,poh2013user}, rank-~\cite{connaughton2013fusion,monwar2009multimodal,liu2017robust}, and decision-level~\cite{kumar2011personal,bhatt2014recognizing,kamyab2016feature} fusion schemes, feature-level fusion results in a better discriminative classifier~\cite{nie2017enhancing, faundez2005data,ross2005feature} due to preservation of raw information~\cite{shekhar2014joint}. Feature-level fusion integrates different features extracted from different modalities into a more abstract and compact feature representation, which can be further used for verification or identification~\cite{eshwarappa2011multimodal,haghighat2016discriminant}. Several frameworks have exploited feature-level fusion for multimodal biometric identification. Among them, serial feature fusion~\cite{liu2001shape}, parallel feature fusion~\cite{yang2003feature},  Canonical Correlation Analysis (CCA)-based feature fusion~\cite{sun2005new}, Joint Sparse Representation Classifier (JSRC)~\cite{shekhar2014joint}, Supervised Multimodal Dictionary Learning
(SMDL)~\cite{bahrampour2016multimodal}, and  Multiset Discriminant Correlation Analysis  (MDCA)~\cite{haghighat2016discriminant} are the most prominent techniques. 

The prevalent feature fusion method in the deep learning literature is feature concatenation, which becomes very inefficient as the dimensionality of the feature space increases~\cite{nagar2012multibiometric,goswami2016group,soleymani2018multi,shi2016rule}. Bilinear feature multiplication~\cite{lin2015bilinear,chowdhury2016one} is effective since all elements of different modalities interact with each other through multiplication. The main issue in bilinear operation is the high dimensionality of its output regarding the cardinality of the inputs. Recently, to overcome this shortcoming, compact bilinear pooling has been proposed~\cite{gao2016compact,delbrouck2017multimodal,fukui2016multimodal,soleymani2018generalized}. This pooling algorithm mimics results close to bilinear pooling while the dimensionality of the embedding space is relatively small. 

\subsection{Multi-sample recognition}
Multi-sample recognition has been recently utilized in recognition frameworks. The authors in~\cite{yang2017neural} have considered a neural aggregation network, in which a set of face images is represented by a vector in the embedding space. In their proposed framework, a CNN block maps each face image into a feature vector in the embedding space. Their aggregation module consists of two blocks. These blocks adaptively aggregate the feature vectors and form a single fixed-sized feature vector to represent a set of face images. Their framework is trained with a classification loss function without direct supervision. They concluded that their proposed framework can learn to differentiate between high-quality and low-quality face images in a set of images by solely minimizing this loss function.
The authors in~\cite{liu2017quality} have considered the problem where a set of face images are aggregated to be presented by a vector in the embedding space. Their proposed network consists of two branches. The first branch constructs a feature vector in the embedding space for each image sample, while the second branch computes the quality score for each image sample. Finally, all of the feature vectors and quality scores in one set are aggregated through the loss function to construct the feature vector in the embedding space and represent the set of face images. 

\subsection{Quality-aware fusion}
The quality of a biometric sample is defined as its suitability for feature extraction, correct recognition, and automated matching with a high confidence score~\cite{singh2019comprehensive}. Multimodal quality-based fusion frameworks give higher weights to the more reliable modalities. On the other hand, fusion algorithms which do not consider the quality of modality samples provide a fixed weighting scheme. Therefore, these frameworks do not present the optimal decision when sample quality varies. The quality-based fusion frameworks should receive an effective set of quality measures. These frameworks should also present an effective fusion mechanism to consider these quality scores from all samples and make an optimal decision~\cite{poh2012unified}. 

One of the very first works considering the quality of the samples in biometric fusion is presented in~\cite{bigun2003multimodal}. In this work, the authors have manually deployed quality measures generated by human experts. The authors in~\cite{poh2010quality} have proposed a framework to minimize cross-device matching performance degradation by device-specific quality-dependent score normalization. In this framework, each device score is normalized independently. To fuse the outputs of different devices, these scores are combined using a naive Bayes approach. A user-quality-based fusion of biometric modalities is proposed in~\cite{kumar2010improving}. This work quantifies the quality of biometric data by using user templates to incorporate the quality of the sensor data in order to generate a more reliable estimate on the matching scores, while a score-level fusion of the matching scores in the multimodal setting is considered.

A unified framework for quality-based fusion from a Bayesian perspective is proposed in~\cite{poh2012unified}. In this work, the authors have investigated feature-based and cluster-based fusion algorithms for their quality-based framework. The authors in QFuse~\cite{bharadwaj2015qfuse} present an adaptive context switching algorithm coupled with online learning to address uncontrolled noisy conditions and scalability. A probabilistic logic to explicitly take uncertainty and trust into consideration is proposed in~\cite{vishi2017new}. The authors in~\cite{khiari2016quality} have proposed a dynamic weighted sum fusion quality metric while combining unimodal scores. This work proposes a single quality metric for each gallery-probe comparison, instead of incorporating the quality of the gallery and probe images separately. The context weighted majority algorithm presented in~\cite{sivasankaran2018context} introduced score-level and decision-level context-aware biometric fusion methods to consider the context in which biometric inputs are acquired.

In contrast to the works mentioned in this section, the framework proposed here provides a multi-sample multimodal framework which benefits from quality-aware fusion. Our framework provides a unimodal representation for each modality considering intra-modality quality of the samples in that modality and aggregates these representations using their inter-modality quality. The proposed framework is trained in a weakly-supervised fashion by minimizing the proposed multimodal separability loss to uniformly spread the centers of class representations in the embedding space. The proposed multimodal network compactness loss regularizes the multimodal network by minimizing the hyperspherical energy for different layers of the network. The performance of the proposed framework is compared with several state-of-the-art methods mentioned in this section.  

\section{Quality-Aware Multimodal Network} 
Here, we describe our methodology to provide a framework for multi-sample multimodal recognition for inputs consisting of different modalities with different quality and varying number of samples per modality. We describe the notations and the proposed quality-aware fusion in Sections \ref{sec:notations} and \ref{sec:quality-aware fusion}, respectively. This fusion framework consists of two quality-aware fusion blocks. Then, as discussed in Section~\ref{sec:multimodallosses}, we present our training criteria consisting of two loss functions. The multimodal separability loss aims to construct an embedding which provides separability of the representations by uniformly distributing the multimodal class centers in the embedding space. Due to over-parametrization, the multimodal networks suffer from a lack of generalization on unseen samples.  Hence, we propose a second loss function as multimodal network compactness loss to improve the generalization of the framework by  minimizing the hyperspherical energy for different layers of the network. The two proposed loss functions are customized for the multimodal multi-sample settings. In Section~\ref{sec:Experiments}, we study the effect of these loss functions on the performance of the proposed framework.
\subsection{Notations}\label{sec:notations}
The following notations are used throughout this paper:
\begin{itemize}
\item Multimodal sample set, $X$, consists of one or a set of a varying number of samples from each modality.
\item We consider $K$ modalities, $N$ training samples, and $M$ training classes in the proposed framework. 
\item $X_k$ represents the set of samples from the $k^{th}$ modality in multimodal sample set $X$.
\item $X_{ki}$ represents one sample from modality $k$ in multimodal sample set $X$.
\item $L$ represents the number of layers in the architecture and $N_j$ represents the number of kernels in the $j^{th}$ layer. 
\item For simplicity, in this section, we use the notation $k$, $1\leq k\leq K$, for the different modalities. However, in the next section, we replace them with the actual modality names.
\end{itemize}

\begin{figure}[t]
\begin{center}
\includegraphics[width=.95\linewidth]{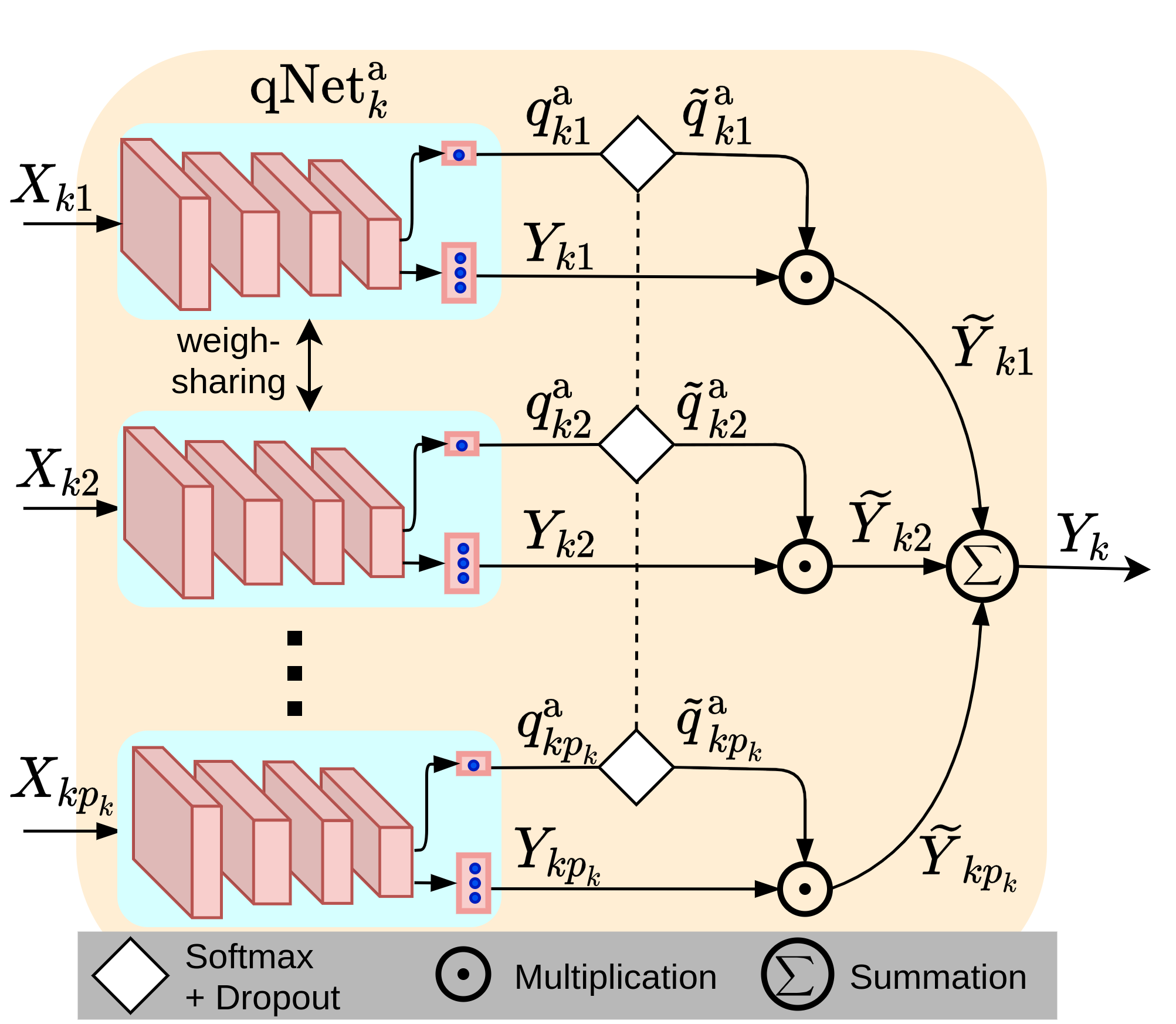}
\end{center}
\caption{$\mathrm{Fusion}_{\,{k}}^{\mathrm{a}}$ consists of a multi-task CNN block, $\mathrm{qNet}_{k}^\mathrm{a}$, and intra-modality aggregation to deliver a unimodal embedding space representation.} 
\label{fig:Att1_2}
\end{figure}

\subsection{Quality-aware fusion mechanism}\label{sec:quality-aware fusion}
We denote each multimodal sample set by $X=\{X_k\}_{k=1}^{K}$, which consists of samples from $K$ modalities, and $X_{k}$ represents samples from the $k^{th}$ modality. Our quality-aware framework consists of two  fusion blocks. The first quality-aware block converts each $X_k$ to a unimodal representation, $Y_k$, while considering the quality of each sample in $X_k$. This fusion block is applied on the samples from each modality and extracts the features that are the best representation of the corresponding modality: $Y_k= \mathrm{Fusion}_{\,{k}}^{\mathrm{a}}(X_k)$. The second quality-aware fusion block constructs the multimodal embedding space representation, ${Z}$, from the unimodal representations of modalities, $Y_k, k=1, 2,...,K$, considering the quality of information across the modalities. This block determines the relative credit assignment to the feature vectors in the unimodal constructed embedding spaces: $Z=\mathrm{Fusion}^{\mathrm{b}}(Y_1,Y_2,...Y_K)$. In the following, we describe each of these fusion blocks in greater detail. 

{\bf Intra-modality fusion ($\mathrm{Fusion}_{\,{k}}^{\mathrm{a}}$):} When multiple samples per modality are available, one can consider the average representation of samples to combine their identification information for the corresponding modality. This is equivalent to assigning equal quality scores to all of these samples. However, a better choice is to incorporate the quality of the samples to combine their representation. Fig. \ref{fig:Att1_2} presents this fusion block consisting of feature extraction, intra-modality quality estimation, and feature aggregation. The inputs to this block are samples from the $k^{th}$ modality, $X_{k1}, X_{k2},...,X_{kp_k}$, where $p_k$ is the number of samples for this modality in the multimodal sample set, and can vary from one sample set to the other. To utilize the quality of samples in $X_k$ and construct a richer unimodal representation, $Y_k$, this block consists of a quality-aware modality dedicated network, $\mathrm{qNet}_{k}^\mathrm{a}$, and the intra-modality feature aggregation.       

The first fusion block aims to provide a discriminative unimodal representation for a set of samples $X_k$. This block constructs a fixed-size vector representation, $Y_{ki}$, and an intra-modality quality score, $q_{ki}$, for sample $X_{ki}$. These representations construct the embedding space representation for the $k^{th}$ modality through softmax normalization of the quality scores: 
\begin{equation}
\widetilde{q}_{ki}^\mathrm{a}=\frac{e^{q_{ki}^\mathrm{a}d_{ki}}}{\sum_j e^{q_{kj}^\mathrm{a}d_{kj}}},
\end{equation}
where $i=1,...,p_k,$ and $\widetilde{q}_{ki}^\mathrm{a}$ represents the normalized intra-modality quality score for the $i^{th}$ sample. These quality scores are utilized to construct the representation of the $k^{th}$ modality in the embedding space, where $q_{ki}^\mathrm{a}Y_{ki}$ represents the normalized quality-aware embedding space vector representation for sample $X_{ki}$: 
\begin{equation}
{Y_{k}}=\sum_i \widetilde{q}_{ki}^\mathrm{a}{Y_{ki}}.
\label{y_k}
\end{equation}
One of the main concerns for the proposed framework is the possibility of high-quality samples dominating the other samples during the training, in which the quality scores corresponding to these samples tend to be significantly high, forcing the scores corresponding to the other samples to be very small. To resolve this issue, we perform the dropout technique on samples of the modality during the training and randomly set some of the quality scores to zero. Here, $d_{ki}$ is a binary value, and takes values based upon the modality-specific dropout probability, $\mu_k$. As presented in Fig.~\ref{fig:Att1_2}, when a modality consists of one sample, $X_{k}=X_{k1}$, the first fusion block acts as a feature extraction block, and provides a discriminative unimodal representation for this sample. 

\begin{figure}[t]
\begin{center}
\includegraphics[width=.95\linewidth]{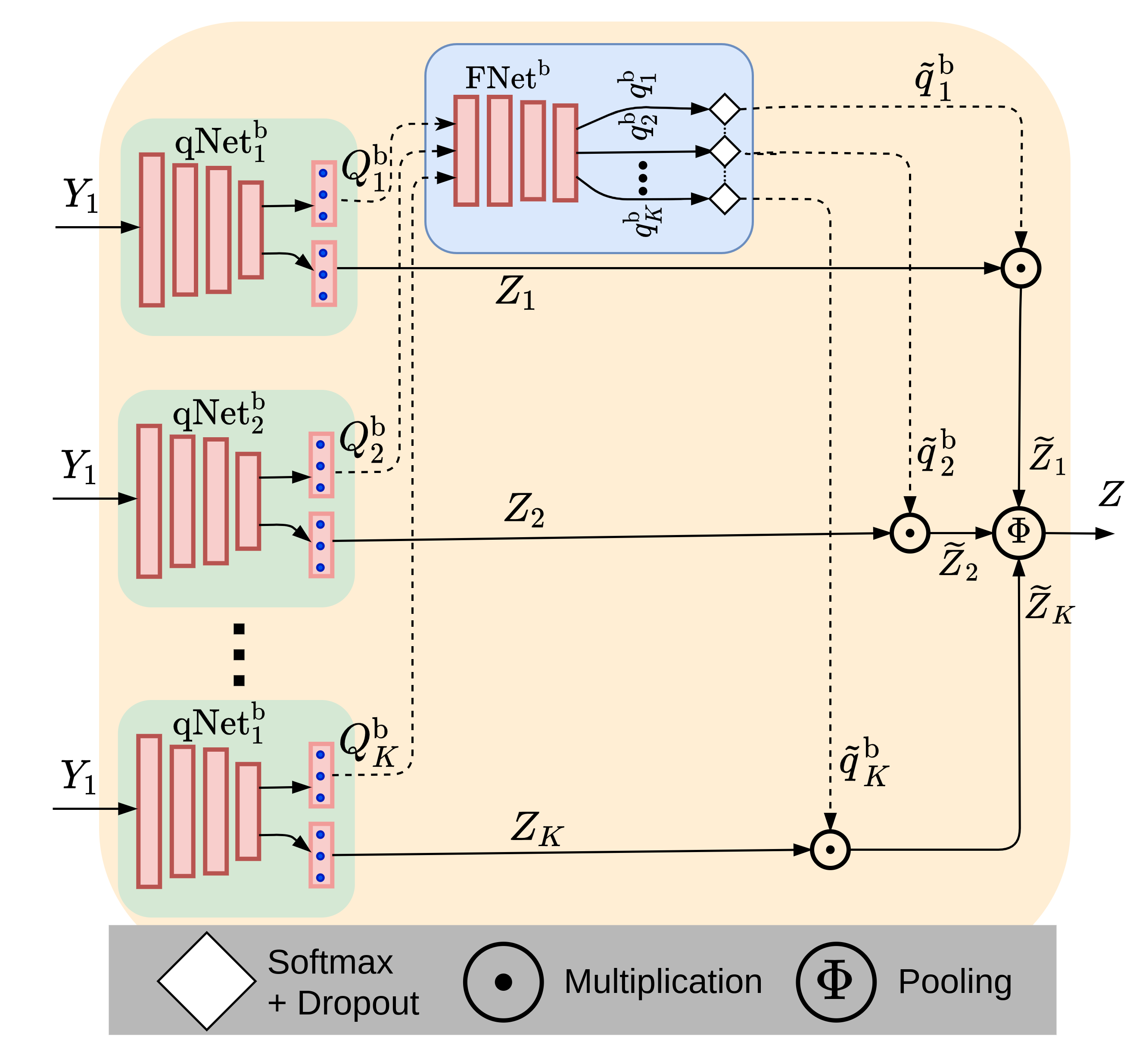}
\end{center}
\caption{$\mathrm{Fusion}_{\,{k}}^{\mathrm{b}}$ consists of $K$ modality-dedicated networks, $\mathrm{qNet}_k^{\mathrm{b}}$, and a fully-connected network, $\mathrm{FNet}^\mathrm{b}$. Each modality-dedicated network, $\mathrm{qNet}_k^{\mathrm{b}}$, presents a modality with a representation and a modality-dedicated quality vector. Quality vectors are concatenated and fed into $\mathrm{FNet}^\mathrm{b}$ to provide the inter-modality quality scores.}
\label{fig:Att1_1}
\end{figure}

{\bf Inter-modality fusion ($\mathrm{Fusion}^{\mathrm{b}}$):} As presented in Fig.~\ref{fig:MDN_0}, the second fusion block consists of feature transformation, inter-modality quality estimation and inter-modality feature aggregation. The feature transformation aims to provide the flexibility for the unimodal representations, $Y_k$, utilized for inter-modality quality estimation and feature aggregation trained by loss functions described in Section~\ref{sec:multimodallosses}. This fusion block includes inter-modality networks, $\mathrm{qNet}_k^{\mathrm{b}}$, and a fully-connected block $\mathrm{FNet}^\mathrm{b}$ to estimate the inter-modality quality scores.\footnote {$\mathrm{qNet}_k^{\mathrm{a}}$ networks have different architectures. However, although parameters for $\mathrm{qNet}_k^{\mathrm{b}}$ networks differ, we consider their architecture to be the same.} This fusion block constructs a multimodal representation for the multimodal sample set, $X$, through learning the inter-modality quality of the unimodal representations. To this aim, the unimodal representations are fed into inter-modality networks ($\mathrm{qNet}_k^{\mathrm{b}}$, $1\leq k\leq K$). This fusion block constructs an embedding space representation, $Z_{k}$, as well as a modality-dedicated quality vector, $Q_{k}^\mathrm{b}$. Each representation and the quality vector, interacting with the corresponding unimodal feature vectors and quality vectors from the other modalities, build $Z$ as the multimodal embedding space representation of $X$. To this aim, as presented in Fig.~\ref{fig:Att1_1}, the quality vectors are concatenated and fed into a fully-connected block of layers, $\mathrm{FNet}^\mathrm{b}$. The quality vectors from all of the modalities interact through this network, and the inter-modality quality scores corresponding to each modality, as $q_k^\mathrm{b}$, $k=1,2,...,K$ are estimated. These quality scores are normalized through softmax normalization: 
\begin{equation}
\widetilde{q}_{k}^\mathrm{b}=\frac{e^{q_{k}^\mathrm{b}d_k}}{\sum_j e^{q_{j}^\mathrm{b}d_j}},\;\; k=1,...,K,
\label{s_j}
\end{equation} 
where $\widetilde{q}_k^\mathrm{b}$ represents the normalized inter-modality quality score for the $k^{th}$ modality in $X$ and present the relative importance of this modality in the recognition of $X$. To avoid the possibility of one modality dominating the other modalities during the training, we randomly set some of these quality scores to zero. This approach is implemented utilizing binary values, $d_k$, which take values based upon the dropout probability, $\mu$. In the case of single-sample multimodal recognition, each normalized inter-modality quality score interprets the quality of the sample as well as the inter-modality quality of the modality. These normalized quality scores interact with the embedding space representations of the modality samples, $Z_{k}$ vectors, to present a quality-aware representation of $X$, where $\widetilde{q}_k^\mathrm{b}{Z_{k}}$ represents the normalized embedding space representation for $X_{k}$. We aggregate the representations of all the modalities as:  

\begin{equation}
Z=\Phi_{k=1}^{K}(\widetilde{q}_k^\mathrm{b} {Z_k}),
\label{z_pool}
\end{equation}
where $\Phi$ represents the aggregation method applied to the normalized embedding space vectors, such as addition~\cite{haghighat2016discriminant}, concatenation~\cite{goswami2016group}, bilinear multiplication~\cite{chowdhury2016one}, or compact bilinear pooling~\cite{gao2016compact}. In all experiments presented in this paper, we consider addition of the feature vectors as the aggregation method, which results in $Z=\sum_{k=1}^{K}\widetilde{q}_k^\mathrm{b} {Z_k}$. Equivalently, this operation represents re-weighting the last layers of the modality-dedicated networks. The learned representations, $Z$ and $Y_k$, are considered for multimodal and unimodal frameworks, respectively. These representations are utilized through the loss functions described in the next section to construct the decision. 

The proposed framework learns inter-modality quality scores by minimizing the recognition loss function. These scores represent both the quality of the samples of one modality assigned to each multimodal sample and the inter-modality quality of the modality compared to the other modalities for a multimodal sample set. Therefore, the inter-modality quality of the $k^{th}$ modality over the dataset for the underlying recognition task is computed as:
\begin{equation}
P_{k}^\mathrm{b}=\mathbb{E}_X\{\widetilde{q}_k^\mathrm{b}\}, 
\label{quality_actual}
\end{equation}
where $\mathbb{E}_X$ represents the expectation over the multimodal sample sets in the dataset. 
\begin{figure}[t]
\begin{center}
    `\includegraphics[width=.8\linewidth]{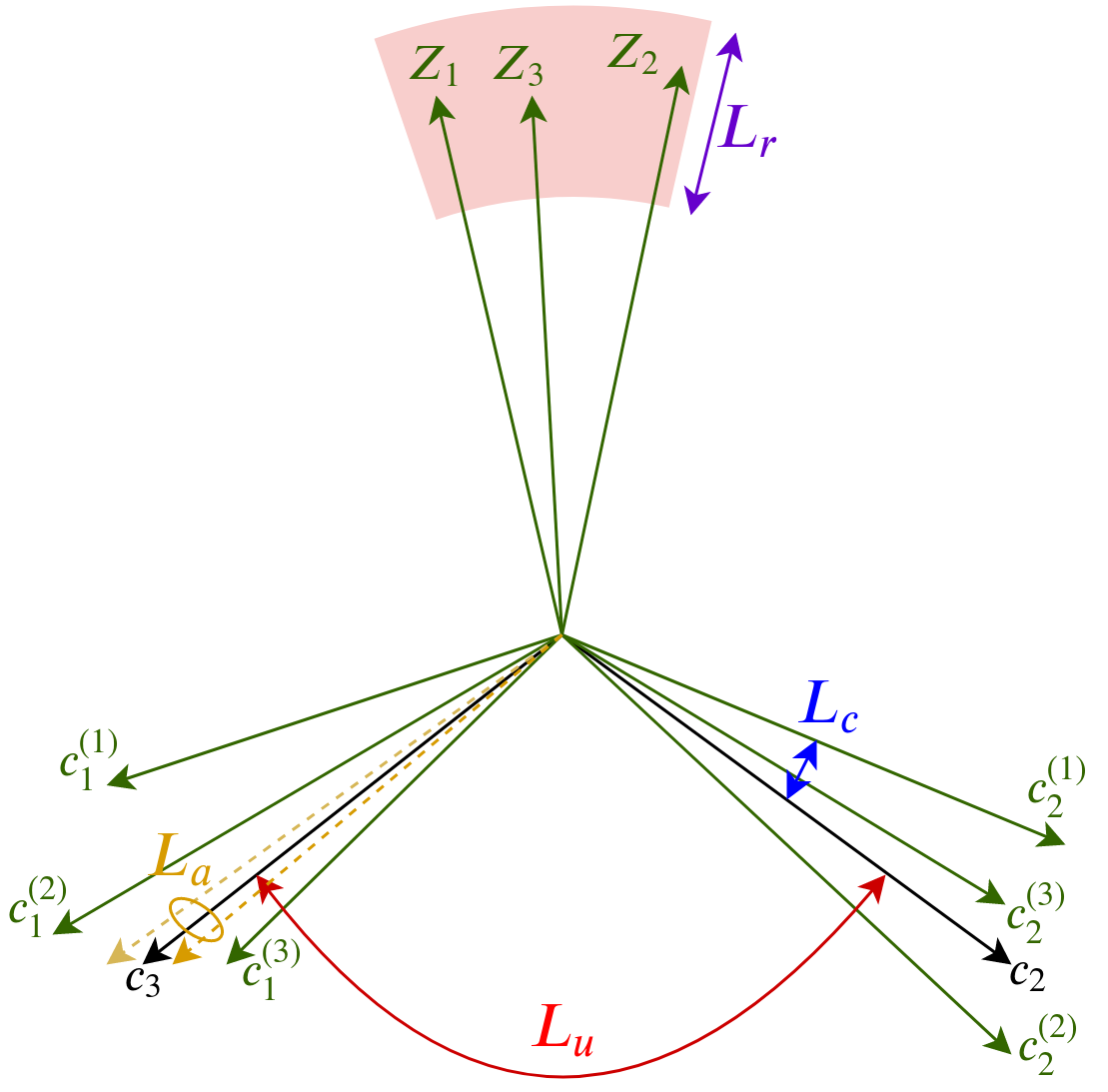}
\end{center}
\caption{Our multimodal separability training loss consists of four multimodal losses. The angular loss, $L_a$, provides the compactness between different multimodal sample sets of a given class. The uniform loss, $L_u$, aims to guarantee that the centers for different multimodal classes are uniformly distributed in the embedding space. $L_c$ forces the centers of different modalities for a class to follow the same direction. For each class, $L_r$ aims to provide $Z_k$ representations that are comparable. }
\label{fig:loss}
\end{figure}
\subsection{Multimodal separability and network compactness}\label{sec:multimodallosses}
Inspired by the recent advances in metric learning for deep biometric recognition such as SphereFace~\cite{liu2017sphereface}, ArcFace~\cite{deng2019arcface}, and UniformFace~\cite{duan2019uniformface}, we design an equi-distributed multimodal embedding in which the inter-class distance of multimodal representations is maximized and the intra-class variance is minimized. The equi-distributed constraint is applied on the centers of each class representations by uniformly spreading them in the embedding space. However, simple consideration of this constraint is not sufficient in a multimodal framework in practice. Thus, we consider two additional constraints on the unimodal embeddings, linking them to the multimodal embedding, which subsequently results in unifying the unimodal representations.   
In addition, to boost the generalization capability of the proposed framework, we regularize the weights in each layer of the architecture. The proposed regularization method benefits from considering similar, but not necessarily the same, architectures for different networks. 

{\bf Separability of representations:} The softmax loss is the common loss used for training CNN-based classifiers in the literature. This loss is defined as a combination of the last fully-connected layer, a softmax function, and a cross-entropy loss~\cite{liu2016large}. The radial properties of features learned by the softmax loss do not contribute to the discrimination of samples, thereby the angular similarity should be preferred, leading to normalized features~\cite{zheng2018ring}. 

Let us assume that $N$ is the number of training samples, $x_i$ is the learned feature representation corresponding to the $i^{th}$ training sample with label $y_i$, and $v_j$ and $b_j$ are the weights and bias of the last fully connected layer corresponding to $j^{th}$ class, respectively. 
To impose the angular similarity to the softmax loss, we assume that $||v_j||=1$ and $b_j=0$. These assumptions result in the classification to depend entirely on the angles between $x_i$ and $v_j$, $\theta_{j, i}$. Therefore, the modified softmax loss can be defined only based on $\theta_{j, i}$~\cite{liu2017sphereface}. Several works have provided more general assumptions on the modified softmax loss for different recognition tasks: 

\begin{equation}
\small
    L_{a}=-\frac{1}{N}\sum_i\log{\frac{e^{||x_i||(\cos(m_1\theta_{y_i,i}+m_2)-m_3)}}{e^{||x_i||(\cos(m_1\theta_{y_i,i}+m_2)-m_3)}+\sum\limits_{j\neq i}{e^{||x_i||\cos(\theta_{j,i})}}}}, 
\label{angualrloss}    
\end{equation}
where $L_a$ represents the angular similarity loss. The effect of $m_1$, $m_2$, and $m_3$ are studied in  SphereFace~\cite{liu2017sphereface}, ArcFace~\cite{deng2019arcface}, and CosFace~\cite{wang2018cosface}, respectively. Inspired by the UniformFace~\cite{duan2019uniformface}, we maintain the discriminative nature of the framework considering the angular loss, while forcing the embedding space representations to follow a uniform distribution. Here, during the training, centers $c_{j_1}$ and $c_{j_2}$ are assigned to the ${j_1}^{th}$ and ${j_2}^{th}$ classes~\cite{duan2019uniformface}. Then, the uniform loss is defined as: 
\begin{equation}
L_{u}=\frac{1}{M(M-1)}\sum_{j_1=1}^{M}\sum_{\substack{j_2=1\\j_2\neq j_1}}^M\frac{1}{||c_{j_1}-c_{j_2}||_2+1}, 
\label{uniformloss}
\end{equation}
where $M$ is the number of classes during the training phase. We combine the uniform loss and angular softmax loss as uniform angular loss:

\begin{equation}
L_{1}= L_a+\lambda_u L_u,
\label{uniformsoftmax}
\end{equation}
where $\lambda_u$ is the regularization parameter. 

As described in Equation~\ref{L_ms}, we train the multimodal representation, $Z$ and each of the unimodal representations, $Y_k$, using Equation~\ref{uniformsoftmax}. However, to enforce the unimodal representations, $Z_k$, for different modalities of a class to be comparable for both estimating the inter-modality quality scores and multimodal fusion, we define a representation loss between these representations. In particular, we want the magnitude and phase of $\widetilde{q}_{k}^\mathrm{b}{Z_{k}}$ to capture the inter-modality quality of the corresponding modality and its recognition information, respectively. Therefore, we constrain the magnitude of the modality representations to depend solely on $\widetilde{q}_{k}^\mathrm{b}$:

\begin{equation}
L_{r}=\frac{1}{NK(K-1)}\sum_{i=1}^N\sum_{k_1=1}^{K}\sum_{\substack{k_2=1\\k_2\neq k_1}}^{K}\frac{{(||Z_{k_1}||_2-||Z_{k_2}||_2)^2}}{\sum_{k=1}^K ||Z_k||_2},
\end{equation}
where the first summation represents multimodal sample sets in the training set and the denominator represents the summation of the norms of $Z_k$ representations for a multimodal sample set.  

We consider that for each class, $Z_k$ representations share the same direction. This assumption provides a better separability between the classes since small variations of unimodal representations for each modality result in a minimal multimodal intra-class variance. Equivalently, the unimodal representations for different modalities of the same class should represent the same directions as the multimodal embedding space representation of that class. We define a similarity loss between the directions of the centers of the modalities of the same class as: 
\begin{equation}
L_{c}=\frac{1}{KM}\sum_{k=1}^{K}\sum_{j=1}^{M}{||\frac{c_{j}}{||c_{j}||_2}-\frac{c_{j}^{(k)}}{||c_{j}^{(k)}||_2}||^2}, 
\end{equation}
where $c_{j}^{(k)}$ represents the representation center for $k^{th}$ modality of the $j^{th}$ class. We define multimodal separability loss as: 
\begin{equation}
\small
L_{ms}= \underbrace {L_a+\lambda_u L_u}_{L_1}+\underbrace{\lambda_c L_c+\lambda_r L_r}_{L_2}+\frac{1}{K}\sum_{k=1}^K (\lambda_{ak} L_{ak}+\lambda_{uk} L_{uk}),
\label{L_ms}
\end{equation}
where $L_{ak}$ and $L_{uk}$ represent the unimodal uniform angular loss function for the $k^{th}$ modality and $\lambda_r$, $\lambda_c$, $\lambda_{ak}$, and $\lambda_{uk}$ are the regularization parameters. In verification setups, $\lambda_c=0$. In this training loss, $L_2$ represents the inter-modality training loss, while $L_1$ represents multimodal uniform angular training loss. Fig.~\ref{fig:loss} highlights the effect of the four defined losses on the separability of our multimodal framework. It is worth mentioning that the last term in the above equation represents the unimodal uniform angular training loss~\cite{duan2019uniformface}. Therefore, while computing it, the unimodal centers of classes are considered.    

{\bf Network compactness:} Although deep neural networks are powerful nonlinear functions that can be trained end-to-end to extract the features and satisfy the underlying recognition task simultaneously, their over-parametrization results in highly correlated neurons that can hurt the generalization ability
and incur unnecessary computation cost~\cite{liu2018learning}. Multimodal deep neural networks suffer from this shortcoming the most, since they require a vast number of parameters and training multimodal sample sets. Regularization of the deep neural networks aims to avoid the representation redundancy. Regularization of these networks can roughly be categorized into implicit and explicit methods~\cite{lin2019compressive}. 

\begin{figure}[t]
\begin{center}
\includegraphics[width=.99\linewidth]{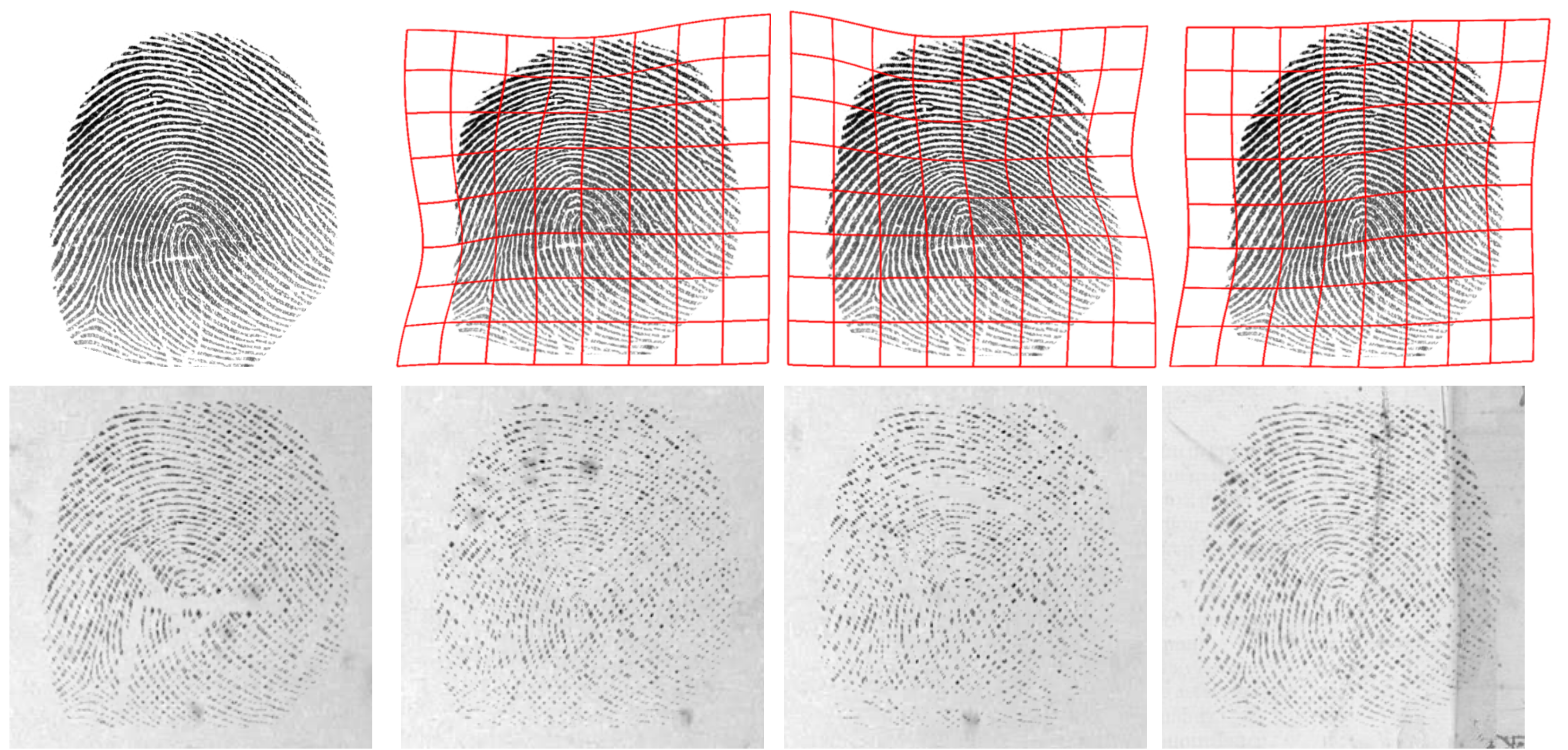}
\end{center}
\caption{Corrupted fingerprint samples from the BioCOP dataset in the training set generated from the clean sample (top left) by warping the clean fingerprint~\cite{dabouei2018fingerprint} (first row) and fading the fingerprint ridges at random points and adding backgrounds~\cite{dabouei2018id} (second row). }
\label{fig:r0}
\end{figure}

Implicit methods do not directly impose constraints on the weights, but instead, regularize the networks in order to prevent over-fitting and stabilize the training dynamics. Batch normalization~\cite{ioffe2015batch}, dropout~\cite{srivastava2014dropout}, weight normalization~\cite{salimans2016weight}, and group normalization~\cite{wu2018group} are examples of the implicit regularization methods. Explicit models, such as orthonormal regularization~\cite{brock2017neural,liu2017deep}, diversification~\cite{xie2016diversity,liu2018decoupled}, uncorrelation~\cite{xie2017uncorrelation,cogswell2016reducing}, and minimizing the hyperspherical energy (MHE)~\cite{liu2018learning}, aim to impose direct constraints on the weights of the network. However, the high-dimensionality of the kernels in convolutional neural networks, in addition to the vast number of kernels in multimodal frameworks, makes it difficult to regularize our multimodal framework using explicit methods~\cite{wang2019makes}. Here, we expand {\it Compressive Hyperspherical Energy Minimization} (CoMHE)~\cite{lin2019compressive}, which projects the kernels and neurons of the network to a low-dimensional space and minimizes the energy in the projected space, to apply it in our multimodal setting. We define the hyperspherical energy for the $j^{th}$ convolutional layer which consists of $N_j$ kernels, ${\bf{W}}_{j}=\{w_1,w_2,...,w_{Nj}\}$ as~\cite{liu2018learning}:   
\begin{equation}
E_s({\bf{W}}_{j})= \sum_{i=1}^{N_j}\sum_{l=1, l\neq i}^{N_j} (||g(\hat{w_i})-g(\hat{w_l})||_2)^{-2},
\label{E_energy}
\end{equation}
where $\hat{w_i}=\frac{w_i}{||w_i||_2}$. However, the proposed energy minimization problem can result in colinear kernels in opposite directions. Therefore,  we consider MHE in half space, in which both $\hat{w_i}$ and $-\hat{w_i}$ are utilized in the energy function above. The same energy function can be applied to the fully-connected layers where the vector $w_i$ represents the weights going to the $i^{th}$ neuron. 

The compression function is defined as $g(\hat{w_i})=\frac{P^{*}\hat{w_i}}{||P^{*}\hat{w_i}||}$, where $P^{*}$ is the optimized projection matrix using unrolled optimization~\cite{lin2019compressive}. Then, the hyperspherical loss can be defined as:

\begin{equation}
\small
L_{mc}=\lambda_h\sum_{j=1}^{L-1}\frac{1}{N_j(N_j-1)}\{E_s\}_j+\lambda_{h_0}\frac{1}{N_L(N_L-1)}E_s({\hat{w}_i}^{out}|_{i=1}^{M}),
\label{Lh}
\end{equation}
where $L$ is the number of layers. 
For our multimodal framework, the $P^*$ matrices for different modalities are shared when the kernel size is the same. We find it beneficial to share projection matrices for different modalities. Sharing the projection basis can effectively reduce the number of projection parameters, also reducing the inconsistency within the hyperspherical energy minimization of projected neurons for different modalities, and further improves the generalization. To implement this loss function, we consider the dimension of the projected space to be equal to $30$ for all layers. For the rest of this paper we refer to this loss function as multimodal network compactness loss, $L_{mc}$. Then, the overall training loss is defined as: 
\begin{equation}
L=L_{ms}+L_{mc}.
\label{Ltrain}
\end{equation}

\begin{figure}[t]
\begin{center}
\includegraphics[width=.99\linewidth]{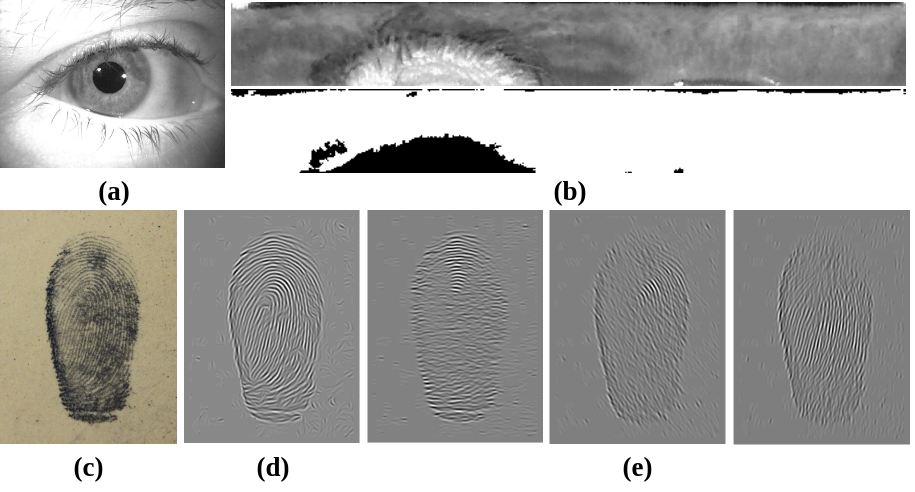}
\end{center}
\caption{ (a) Eye image, (b) normalized iris and mask images, (c) latent fingerprint image, (d) enhanced fingerprint image using~\cite{jain2000filterbank}, and (e) three enhanced fingerprint images using constant Gabor angles of $0^\circ, 60^\circ$, and $100^\circ$ for the whole image. }
\label{fig:preprocess}
\end{figure}

\begin{figure*}[t]
\begin{center}
\includegraphics[width=.99\linewidth]{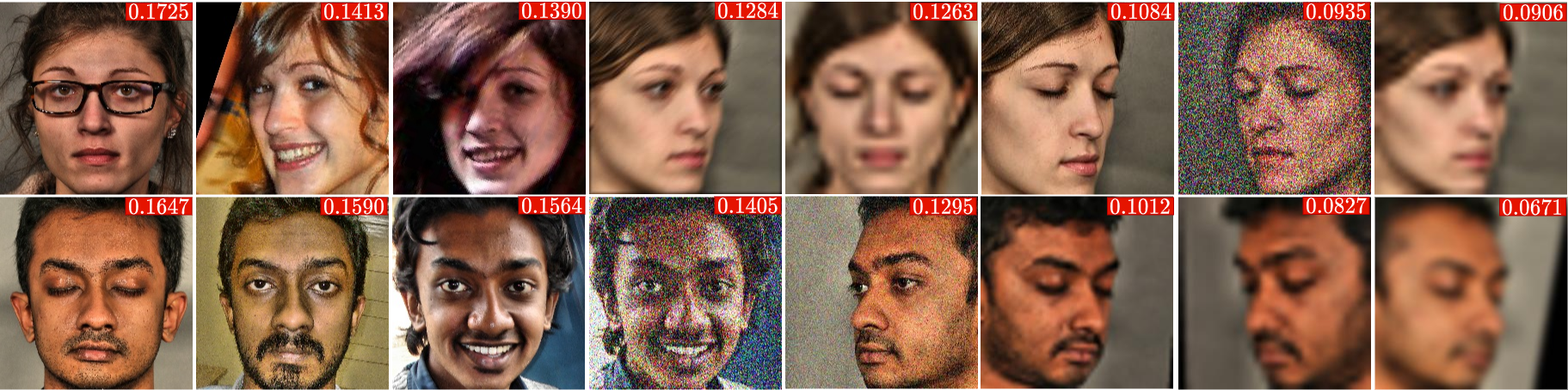}
\end{center}
\caption{The estimated quality of the corrupted samples from the BioCOP dataset in a multi-sample unimodal framework.}
\label{fig:r1}
\end{figure*}

\section{Experiments and discussions}\label{sec:Experiments}
In this section, we present performance metrics, the data representation for different modalities, training setup, experimental scenarios, and results. We conclude this section with the discussions and comparisons with the state-of-the-art classical and deep learning algorithms to address multi-sample multimodal recognition problem. 
To evaluate the performance of the proposed framework, we follow the evaluation metrics and protocols presented in~\cite{klare2015pushing}.
For the identification setup, the \emph{Recall} metric is used. This metric computes the probability that a subject is correctly classified at least at the specified rank, while the candidate classes are sorted by their similarity score to the query samples. The performance metrics for the verification setup are the area under the curve (AUC), equal error rate (EER), and true acceptance rate (TAR) at different false acceptance rates (FAR). In addition, cumulative match curve (CMC) and receiver operating characteristic (ROC) are considered to present the performance for identification and verification setups, respectively.

{\bf Data representation:} To preprocess the samples, the face images are aligned through five landmarks (two eyes, two mouth corners and nose)~\cite{zhang2016joint}, and cropped to $112\times 112$ resolution images. As presented in Fig.~\ref{fig:preprocess}(b), iris images are segmented, normalized using OSIRIS~\cite{othman2016osiris}, and transformed into $64\times 512$ strips. In addition, each iris image is concatenated in depth with its mask image. Fingerprint images are enhanced using the method described in~\cite{chikkerur2004systematic}, in which the core point is detected from the enhanced image~\cite{jain2000filterbank}, and a $224 \times 224$ region centered by the core point is cropped for recognition. We follow the conventional Gabor filtering for enhancing fingerprint ridge information~\cite{jain2000filterbank}. 

This approach identifies the locally optimal Gabor filter using the estimated ridge frequency and orientation maps. However, since in our problem fingerprints are assumed to be of different quality, these maps and the subsequent filtering can be significantly deteriorated. Hence, instead of estimating the best local Gabor filter, which is unreliable for low-quality samples, we feed the network with the response of several major Gabor filters with varying angles. We assume that the network learns to select the appropriate response through minimizing the recognition loss. 
Each fingerprint image is concatenated in depth with nine other images. The algorithm described in~\cite{jain2000filterbank} computes the direction of the Gabor filter to estimate the ridge maps locally. Each of these additional nine images is the response of Gabor filtering with the angels in $[0^\circ\!-\!160^\circ]$ with steps of $20^\circ$. Figs~\ref{fig:preprocess}(d) and~\ref{fig:preprocess}(e) visualize several Gabor responses obtained from the latent fingerprint in Fig.~\ref{fig:preprocess}(c).

\subsection{Training setup}
{\bf Training datasets:} The BioCOP multimodal dataset{\footnote {This dataset is available upon request: Jeremy.Dawson@mail.wvu.edu.}} is one of the few datasets that allows training of multi-sample multimodal fusion since it contains a vast number of samples from different modalities from the same individual. This dataset consists of four sub-collections acquired over the course of 5 years, labeled by the year when each sub-collection was initiated; 2008, 2009, 2012, and 2013. There are 3,990 distinct subjects in these four sub-collections, while there are subjects common in these sub-collections, e.g., 294 in sub-collections 2012 and 2013.  We consider face, iris, and fingerprint samples from this dataset in our training phase. There are a total number of 254,660, 264,821, and 338,912 samples for face, iris, and fingerprint modalities in this dataset, respectively.

The face modality contains both constrained and unconstrained images with different expressions, camera angles, and camera models. The constrained face images are acquired in head pose angles of $\pm 90^\circ$, $\pm 45^\circ$, and $0^\circ$, with open and closed eyes. The fingerprint modality consists of all ten fingers captured using the {\it CrossMatch Verifier 300LC}, {\it CrossMatch Verifier 310}, and {\it UPEK EikonTouch 700} sensors. The iris samples contain both left and right irises and are acquired using the {\it Aoptix Insight}, {\it CrossMatch I SCAN 2}, and {\it LG ICAM 4000} near-infra-red sensors. Although, the interval of data acquisition in BioCOP dataset can vary up to five years, we also utilize the VGGFace2~\cite{cao2018vggface2} dataset to consider age-progression during the training phase. The VGGFace2 dataset also provides the training setup with more pose variations. This dataset consists of 9,131 subjects with 3.31 million face images. These images include different pose, quality, and resolution variations. As described in the {\bf Training} section in more details, for half of the subjects which have the least number of face samples, the face samples in BioCOP dataset are replaced with face samples from the VGGFace2 dataset.

\begin{table}[t]
\caption[Table caption text]{The input size and network architectures. The first row for each network represents the main branch which delivers the embedding space representation and the second row represents the quality score branch.} 
\scriptsize 
\begin{center}
\addtolength{\tabcolsep}{-5pt}
\begin{tabular}{l@{\hskip .05in}l@{\hskip .05in}l@{\hskip .05in}l@{\hskip .05in}}
\toprule 
network & input & architecture\\
\hline
\multirow{2}{*}{${\mathrm{qNet}_\mathrm{Face}^\mathrm{a}}$}&\multirow{2}{*}{$112\!\times\!112\!\times\!3$}& \multirow{2}{*}{C64-$3\!\times\!$RES64-$3\!\times\!$RES128}&-$3\!\times\!$RES256-FC512\\
&&$\qquad\qquad\qquad\qquad\qquad\;\;$&-M-FC1\\
\hline
\multirow{2}{*}{${\mathrm{qNet}_\mathrm{Iris}^\mathrm{a}}$}&\multirow{2}{*}{$64\!\times\!512\!\times\!2$}& \multirow{2}{*}{$2\!\times\!$C64-$3\!\times\!$RES64-$3\!\times\!$RES128}&-$3\!\times\!$RES256-FC512\\
&&$\qquad\qquad\qquad\qquad\qquad\;\;\;\;\;$&-M-FC1\\
\hline
\multirow{2}{*}{${\mathrm{qNet}_\mathrm{Fing}^\mathrm{a}}$}&\multirow{2}{*}{$224\!\times\!224\!\times\!10$}& \multirow{2}{*}{$2\!\times\!$C64-$3\!\times\!$RES64-$3\!\times\!$RES128}&-$3\!\times\!$RES256-FC512\\
&&$\qquad\qquad\qquad\qquad\qquad\;\;\;\;\;$&-M-FC1\\
\toprule 
\hline
\multirow{2}{*}{$\mathrm{qNet}_k^{\mathrm{b}}$}&\multirow{2}{*}{$512\!\times\!1\!\times\!1$}&\multirow{2}{*}{FC512}&-FC512\\
&&$\qquad\;\;\;$&-FC64-FC1\\
\hline
\hline
{$\mathrm {FNet}^\mathrm{b}$}&$16K\!\times\!1\!\times\!1$&FC16-FC16-FC$\!K$\\
\bottomrule
\end{tabular}
\end{center}
\label{table:architecture}
\end{table}

\begin{figure*}[t]
\begin{center}
\includegraphics[width=.99\linewidth]{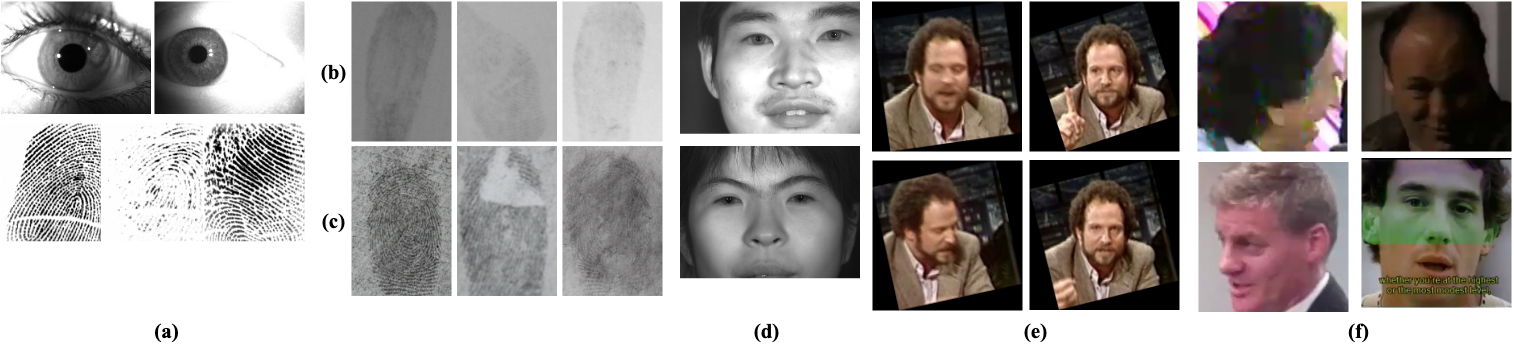}
\end{center}
\caption{Samples from test datasets before preprocessing: (a) BIOMDATA non-ideal, (b) IIIT-Delhi MOLF-Latent (D4), (c) IIIT-Delhi Latent, (d) CASIA Iris V4-Distance, (e) YouTube Face, and (f) IJB-A.}
\label{fig:r2}
\end{figure*}

\begin{table*}[t]
\footnotesize
\caption[Table caption text]{Test datasets description, fine-tuning of the test datasets, and training parameters. For the verification setups, $\lambda_c=0$.} 
\begin{center}
\addtolength{\tabcolsep}{-5pt}
\begin{tabular}{l@{\hskip .05in}c@{\hskip .05in}l@{\hskip .05in}l@{\hskip .05in}c@{\hskip .05in}c@{\hskip .05in}c@{\hskip .05in}c|@{\hskip .05in}c@{\hskip .05in}c@{\hskip .05in}c@{\hskip .05in}c@{\hskip .05in}l@{\hskip .05in}}
\toprule 
&&&& \multicolumn{4}{c}{Unimodal}&\multicolumn{5}{c}{Multimodal}\\ 
&$\#$ classes&Modalities&Fine-tuning Datasets&$m_1,m_2,m_3$&$\lambda_h,\lambda_{h_0}$&$\lambda_{ak},\lambda_{uk}$&$\mu_{k}$&$m_1,m_2,m_3$&$\lambda_h,\lambda_{h_0}$&$\lambda_{u}$&$\lambda_r,\lambda_c$&$\mu$\\
\hline

\multirow{3}{*}{i}& \multirow{3}{*}{219}
&  BIOMDATA L iris&BIOMDATA R Iris              &$1.2,0.3,0.2$&1.5,1&0.3,0.3&0.1&\multirow{3}{*}{1.1,0.4,0.2}& \multirow{3}{*}{2.5,1}&\multirow{3}{*}{1}&\multirow{3}{*}{0.2,0.2}&\multirow{3}{*}{0.2}\\ 
&& BIOMDATA L index& BIOMDATA R Index           &$1.2,0.4,0.2$&1.5,1&0.3,0.3&0.1& &\\
&& BIOMDATA L thumb&BIOMDATA R Thumb            &$1.2,0.4,0.2$&1.5,1&0.3,0.3&0.1& & \\ 
\hline
\multirow{2}{*}{ii}& \multirow{2}{*}{500}
& IJB-A&VGGFACE2                                &$1.35,0.4,0.15$&1.5,1&0.4,0.4&0.2&\multirow{2}{*}{1.1,0.4,0.2}& \multirow{2}{*}{2,1}&\multirow{2}{*}{1}&\multirow{2}{*}{0.2,0.2}&\multirow{2}{*}{0.3}\\ 
&& MOLF-Latent& MOLF-D4 (remain. cla.)          &$1.2,0.4,0.2$&1.5,1&0.4,0.4&0.1& &\\
\hline
\multirow{3}{*}{iii}& \multirow{3}{*}{142}
& YouTube Face&VGGFACE2                         &$1.35,0.4,0.15$&1.5,1&0.3,0.3&0.2&\multirow{3}{*}{1.1,0.4,0.2}& \multirow{3}{*}{2.5,1}&\multirow{3}{*}{1}&\multirow{3}{*}{0.2,0}&\multirow{3}{*}{0.3}\\ 
&&CASIA-distance L iris& CASIA-distance R iris  &$1.2,0.3,0.2$&1.5,1&0.3,0.3&0.1& & \\
&&IIIT-Dehli Latent& IIIT-Lat. (remain. cla.)   &$1.2,0.4,0.2$&1.5,1&0.3,0.3&0.1& & \\

\hline
\end{tabular}
\end{center}
\label{table:testdatasets}
\end{table*}

To provide real-world scenarios for our training setup, we augment the described datasets with corruptions that reduce image quality. The face images are corrupted using motion blur, JPEG compression, additive Gaussian noise, scaling (width to height ratio $\sim\! 0.9\!-\!1.1$), down-sampling and smoothing. To corrupt the iris images, blurring matrix, $B$, warping, $W$, downsampling, $D$, and additive noise $\bar{n}$ are considered: $\bar{X}=DBWX+\bar{n}$ as described in~\cite{alonso2016very}.  

Similarly, as presented in Fig.~\ref{fig:r0}, the fingerprint images are degraded using two corruptions~\cite{dabouei2018id,dabouei2018fingerprint}. The first corruption consists of warping the clean fingerprints~\cite{dabouei2018fingerprint} by randomly sampling the first two principal warp components extracted from the Tsinghua Distorted Fingerprint Database~\cite{si2015detection,dabouei2018fingerprint}. The other corruption considers fading the fingerprint ridges at random points~\cite{dabouei2018id}. Data augmentation is also performed on the fingerprint images, where 20 samples are generated for each fingerprint image by translating the core point both vertically and horizontally using distances coming from Gaussian distributions~\cite{soleymani2018multi}. Here, ten translated images are generated using a Gaussian distribution with parameters $\mu=0$ and $\sigma=2.5$. The remaining ten augmented images are generated with $\mu=0$ and $\sigma=5$. 

{\bf Architecture:}
As presented in Fig.~\ref{fig:Att1_2}, the main architecture of each $\mathrm{qNet}_k^{\mathrm{a}}$ is a multi-task CNN that delivers a unimodal embedding space representation and a scalar quality score. This architecture consists of a ResNet network~\cite{he2016deep} and a fully-connected modality-dedicated embedding layer of size $512$ to deliver $Y_{ki}$. The quality estimation branch of this network delivers a scalar quality score, $q_{ki}$. As presented in Fig.~\ref{fig:Att1_1}, the $\mathrm{qNet}_k^{\mathrm{b}}$ networks are also multi-task networks consisting of fully-connected layers which deliver an embedding space representation, $Z_k$, of size $512$ and a modality-dedicated quality vector, $Q^\mathrm{b}_k$. The score vectors dedicated to modalities are concatenated and fed into $\mathrm{FNet}^\mathrm{b}$ to estimate the inter-modality quality scores for each modality. 

Table~\ref{table:architecture} lists the architectures for these networks. We use (M) as max-pooling of size $2\times 2$ with stride 2, (C[$i$]) as convolutional layers with $i$ kernels of spatial size $3\times 3$ followed by an M, and (FC[$i$]) as fully-connected layers with $i$ nodes. Element $j\times$RES[$i$] consists of $2j$ residual blocks with skip connections after two convolutional layers with $i$ kernels followed by an M. For each $\mathrm{qNet}_k^{\mathrm{a}}$, the quality estimation branch diverges from the main branch at RES128 and delivers the quality score. This branch contains M and FC layers. ReLU is used as the non-linearity after each layer for all networks, except for the final layer of score estimation branches for the $\mathrm{qNet}_k^{\mathrm{a}}$ networks and $\mathrm{FNet}^\mathrm{b}$ network where $sigmoid$ function is considered to limit the scores in the range [0,1].

{\bf Training:} We initially train each $\mathrm{qNet}_k^{\mathrm{a}}$ for the classification setup with a varying number of modality samples per multimodal sample set, where a feature vector of size $512$ is trained using uniform angular loss and network compactness loss as defined in Equations~\ref{uniformsoftmax} and~\ref{Lh}, respectively. Iris and fingerprint unimodal networks are trained on their respective BioCOP modalities, while the face network is trained on the combination of BioCOP and VGGFace2 datasets. The estimated normalized quality scores for degraded samples in the BioCOP dataset can be found in Fig.~\ref{fig:r1}. Each row in this figure presents eight samples of the same subject to construct the unimodal multi-sample set. The number of samples from a modality in a multimodal sample set is chosen to represent the test datasets. Therefore, up to $30$ samples are considered for the face modality, while for the other two modalities up to five samples are considered. 

As described in Table~\ref{table:testdatasets}, each multimodal network, is trained for the multimodal setup, while the multimodal separability loss and multimodal network compactness losses are enforced. This setup is trained for 3,990 subjects in the BioCop dataset, where, for half of the subjects which have the least number of face samples, the face samples are replaced with face samples from the VGGFace2 dataset. To study the effect of data augmentation on the $\mathrm{qNet}^{\mathrm{a}}_\mathrm{Fing}$, we compare the rank-25 recognition rate, with and without data augmentation, with NBIS software~\cite{ko2007users}. Data augmentation improves the performance of the proposed framework from $14.29\%$ to $17.87\%$, while the NBIS software results in $12.72\%$. 

The main branch of $\mathrm{qNet}_k^{\mathrm{a}}$ networks are initialized with weights pre-trained on Imagenet~\cite{deng2009imagenet}. The other parameters are initialized using Kaiming initialization~\cite{he2015delving}. The preprocessing algorithm consists of the channel-wise mean subtraction. The five-fold cross-validation method is considered to estimate the best hyperparameters during the training phase. The training algorithm is deployed using mini-batch stochastic gradient descent with momentum of $0.9$. The training is regularized by weight decay of $5\times 10^{-4}$ and $50\%$ dropout for the fully-connected layers, except for the last layer of each network where the representations are considered for recognition. The moving average decay is set to $0.99$ for all the networks except the iris modality, for which it is set to $0.9$. Batch size is set to 32 and 16 for unimodal and multimodal frameworks, respectively. The initial learning rate is set to $0.1$. The learning rate decreases exponentially by a factor of $0.1$ after $10^5$ iterations, and then every $5\times10^4$ iterations, with the final learning rate of $10^{-6}$. 

\subsection{Results}

{\bf Datasets:} 
In our experiments, we consider multimodal dataset BIOMDATA~\cite{crihalmeanu2007protocol}, face datasets  IJB-A~\cite{klare2015pushing} and YouTube  Face (YTF)~\cite{wolf2011face}, iris dataset  CASIA-Distance~\cite{CasiaIrisV4}, and fingerprint datasets IIIT-Delhi MOLF-Latent (D4)~\cite{sankaran2015multisensor} and IIIT-Delhi Latent fingerprint (D4)~\cite{sankaran2011matching}. Samples from these datasets are presented in Fig.~\ref{fig:r2}. To evaluate the performance of the proposed framework, we consider three multimodal datasets. In the first dataset, which can represent a traditional biometric framework, the recognition framework has access to only one sample from each modality. The second experimental scenario characterizes a multi-biometric identification framework where multiple samples extracted from a low-quality video footage and a varying number of latent fingerprints are available. The third experimental scenario, which can model a access control security system, studies the possibility of improved recognition when multiple samples are available for multiple modalities, e.g., face, iris and fingerprint. To evaluate the performance of the proposed framework for these scenarios, we consider three datasets corresponding to these three scenarios.

There are few multimodal datasets captured in real-world circumstances where each modality consists of multiple samples. Therefore, in our multimodal test setup, except for the BIOMDATA multimodal dataset, we create virtual subjects by assigning real-world biometric samples from subjects in one dataset to the subjects in other dataset {\it i.e.}, chimeric pairing. For instance, we consider the face samples from IJB-A dataset and fingerprint samples from IIIT-Dehli MOLF fingerprint dataset to create our second multimodal dataset. It might be noted that this procedure is feasible since modalities considered in this work are intrinsically independent~\cite{velardo2012improving,lopes2019chimerical,dass2005principled}. For each dataset, the number of samples per subject and per modality may vary. Therefore, for each subject, up to $25$ multimodal sample sets are randomly constructed. A brief description of each multimodal dataset, the fine-tuning, and the hyper-parameters for fine-tuning the architecture for each test dataset are presented in~Table~\ref{table:testdatasets}. It is worth mentioning that, although virtual subjects inherently can provide different recognition performances because of the quality of the samples assigned to them e.g., thumb compared to index fingerprints, the same virtual subjects are considered for all the baselines, which results in a fair comparison. In addition, to provide a better performance assessment for chimeric datasets, we expanded our experiments by evaluating the standard deviation of the multimodal recognition performance over five different sets of virtual subjects from the unimodal datasets. 

\begin{table}[t]
\caption[Table caption text]{The performance for the BIOMDATA non-ideal multimodal dataset.}
\begin{center}
\begin{tabular}{l@{\hskip .05in}c@{\hskip .05in}c@{\hskip .05in}c@{\hskip .05in}}
\toprule 

Method &  @$10^{-4}$ & Rank-1 &EER \\ 
\toprule 
JSRC~\cite{shekhar2014joint}                                &27.13&97.15&7.12\\
GJSRC~\cite{primorac2018generalized}                        &28.54&97.24&6.84\\
SMDL~\cite{bahrampour2016multimodal}                        &28.14&97.12&6.15\\
MDCA~\cite{haghighat2016discriminant}                       &30.41&98.51&5.94\\
VeriFinger+OSIRIS-Sum                                       &30.85&98.28&5.87\\
VeriFinger+OSIRIS-Major                                     &29.54&97.86&6.15\\
CNN-Major                                                   &32.96&99.12&5.65         \\
CNN-Sum                                                     &38.33&99.22&5.21         \\
Weighted feature fusion~\cite{soleymani2018multi}           &43.83&99.40&4.84         \\
Multi-abstract fusion~\cite{soleymani2018multi}             &54.06&99.57&2.97         \\
Generalized compact bilinear~\cite{soleymani2018generalized}&58.30&99.74&2.45         \\
Ours w/o $L_2$                                              &89.06&99.80&0.86\\
Ours w/o $L_c$                                              &{\bf90.13}&99.84&{\bf0.82}\\
Ours with weight sharing                                     &86.82&99.80&0.90\\
Ours                                                        &90.11&{\bf99.92}&{\bf0.82}\\
\hline
\end{tabular}
\end{center}
\label{table:results_dataset1}
\end{table}

{\bf Baselines:} Unimodal matching algorithms considered as baselines for the iris modality are OSIRIS (Version 4.1)~\cite{othman2016osiris}, Sun {\it et al.}~\cite{sun2008ordinal}, and Zhao {\it et al.}~\cite{zhao2017towards}. The performance of the fingerprint modality is compared to NBIS (Release 5.0.0)~\cite{ko2007users} and VeriFinger (Version 10.0)~\cite{VeriFinger}. In addition, angular decision-making algorithms such as SphereFace~\cite{liu2017sphereface} and {UniformFace}~\cite{duan2019uniformface} as well as aggregation algorithms such as Neural Aggregation Network (NAN)~\cite{yang2017neural} are considered to build baselines for face recognition performance. The performance of the proposed framework is compared with the decision-level and score-level fusion of the mentioned algorithms as well as the unimodal performance of our proposed framework. To achieve score-level and decision-level fusion we train independent classifiers for each modality. Then, we aggregate the outputs by adding the corresponding scores of each modality or using the majority voting among the independent decisions. These approaches are abbreviated with {\it Sum} and {\it Major}, respectively~\cite{bahrampour2016multimodal}. We also consider the element-wise averaging of the feature vectors representing samples, equivalent to assigning similar quality scores to all the samples in one modality. This approach is abbreviated as {\it Avg}. The performance of the proposed framework is compared with the performance of Joint Sparse Representation Classifier (JSRC)~\cite{shekhar2014joint}, Generalized Joint Sparse Representation Classifier (GJSRC)~\cite{primorac2018generalized}, Supervised Multimodal Dictionary Learning
(SMDL)~\cite{bahrampour2016multimodal}, and Multiset Discriminant Correlation Analysis  (MDCA)~\cite{haghighat2016discriminant}. Multi-abstract fusion and generalized compact bilinear fusion are adopted from \cite{soleymani2018multi} and~\cite{soleymani2018generalized}, respectively. 

\begin{table}[t]
\caption[Table caption text]{The performance for the second multimodal dataset for varying number of latent fingerprint samples per multimodal sample set.}
\begin{center}
\begin{tabular}{l@{\hskip .05in}l@{\hskip .05in}c@{\hskip .05in}c@{\hskip .05in}|c@{\hskip .05in}c@{\hskip .05in}c@{\hskip .05in}}
\toprule 
&&\multicolumn{2}{c}{Verification}&\multicolumn{3}{c}{Identification}\\
& & @$10^{-2}$ & @$10^{-3}$ & Rank-1 &@$10^{-2}$&@$10^{-1}$\\ \hline

                                                    &1&$96.31$&$92.35$    &$96.12$&     $91.64$&$94.57$\\
                                                    &2&$96.74$&$92.59$    &$96.57$&     $92.06$&$94.76$\\
                                                    &3&$97.12$&$93.42$    &$97.67$&     $92.54$&$95.14$\\
\multirow{-4}{*}{\rotatebox[origin=c]{90}{VeriFinger}~\rotatebox[origin=c]{90}{+($\mathrm{qNet}^{\mathrm{a}}$}~\rotatebox[origin=c]{90}{-Face)}~\rotatebox[origin=c]{90}{-Sum}} &4&$97.64$&$94.12$    &$97.93$&     $92.75$&$95.78$\\\hline

                                                    &1&$97.64$&$93.48$    &$98.37$&     $92.97$&$96.48$\\
                                                    &2&$97.88$&$93.72$    &$98.39$&     $93.12$&$96.55$\\
                                                    &3&$97.93$&$94.01$    &$98.41$&     $93.21$&$96.58$\\
\multirow{-4}{*}{\rotatebox[origin=c]{90}{Weighted}~\rotatebox[origin=c]{90}{feature}~\rotatebox[origin=c]{90}{fusion}~\rotatebox[origin=c]{90}{\cite{soleymani2018multi}}} &4&$98.01$&$94.12$    &$98.42$&     $93.25$&$96.61$\\\hline

                                                    &1&$97.67$&$93.52$    &$98.38$&     $93.02$&$96.51$\\
                                                    &2&$98.90$&$93.75$    &$98.40$&     $93.14$&$96.57$\\
                                                    &3&$97.92$&$94.07$    &$98.42$&     $93.21$&$96.59$\\
\multirow{-4}{*}{\rotatebox[origin=c]{90}{General.}~\rotatebox[origin=c]{90}{compact}~\rotatebox[origin=c]{90}{bilinear}~\rotatebox[origin=c]{90}{\cite{soleymani2018generalized}}}&4&$98.02$&$94.18$    &$98.43$&     $93.25$&$96.60$\\
\hline

                                                    &0&$97.34$&$93.14$    &$98.37$&     $92.71$&$96.36$\\
                                                    &1&$97.92$&$94.22$    &$98.48$&     $93.22$&$96.64$\\
                                                    &2&$98.16$&$94.72$    &$98.53$&     $93.43$&$96.75$\\
                                                    &3&$98.31$&$95.03$    &$98.56$&     $93.52$&$96.79$\\
\multirow{-5}{*}{\rotatebox[origin=c]{90}{Ours} }   &4&{\bf98.40}&{\bf95.18}&{\bf98.57}&{\bf93.54}&{\bf96.81}\\
\hline
\end{tabular}
\end{center}
\label{table:dataset2_table}
\end{table}

{\bf First multimodal dataset:} BIOMDATA non-ideal multimodal database-Release 1~\cite{crihalmeanu2007protocol} is a challenging dataset, since many of the samples are damaged with blur, occlusion, sensor noise and shadows~\cite{haghighat2016discriminant}. Six biometric modalities are considered in our experiments: left and right irises, and thumb and index fingerprints from both hands. Our experiments are conducted on 219 subjects that have samples in all six modalities. Following the protocol in~\cite{wang2019towards}, we fine-tune the network on the right index and thumb fingerprints and the right iris samples. The performance of the framework is tested on the left thumb and index fingerprints and the left iris samples as three modalities constructing the multimodal sample sets. In this dataset, there are 1458, 1519 and 1504 images for left iris, left thumb, and left index, respectively.  In the identification setup, for each modality, four randomly chosen samples are used as the gallery and the remaining samples are used for the test set. For any modality in which the number of the samples is less than five, one sample is used as the probe and the remaining samples are used as the gallery. Then, the multimodal sample sets are  generated as described in the {\bf Datasets} section. In the verification setup, the pairs of multimodal sample sets are generated from these described disjoint sub-sets. 

\begin{figure}[t]
\begin{center}
\includegraphics[width=.95\linewidth]{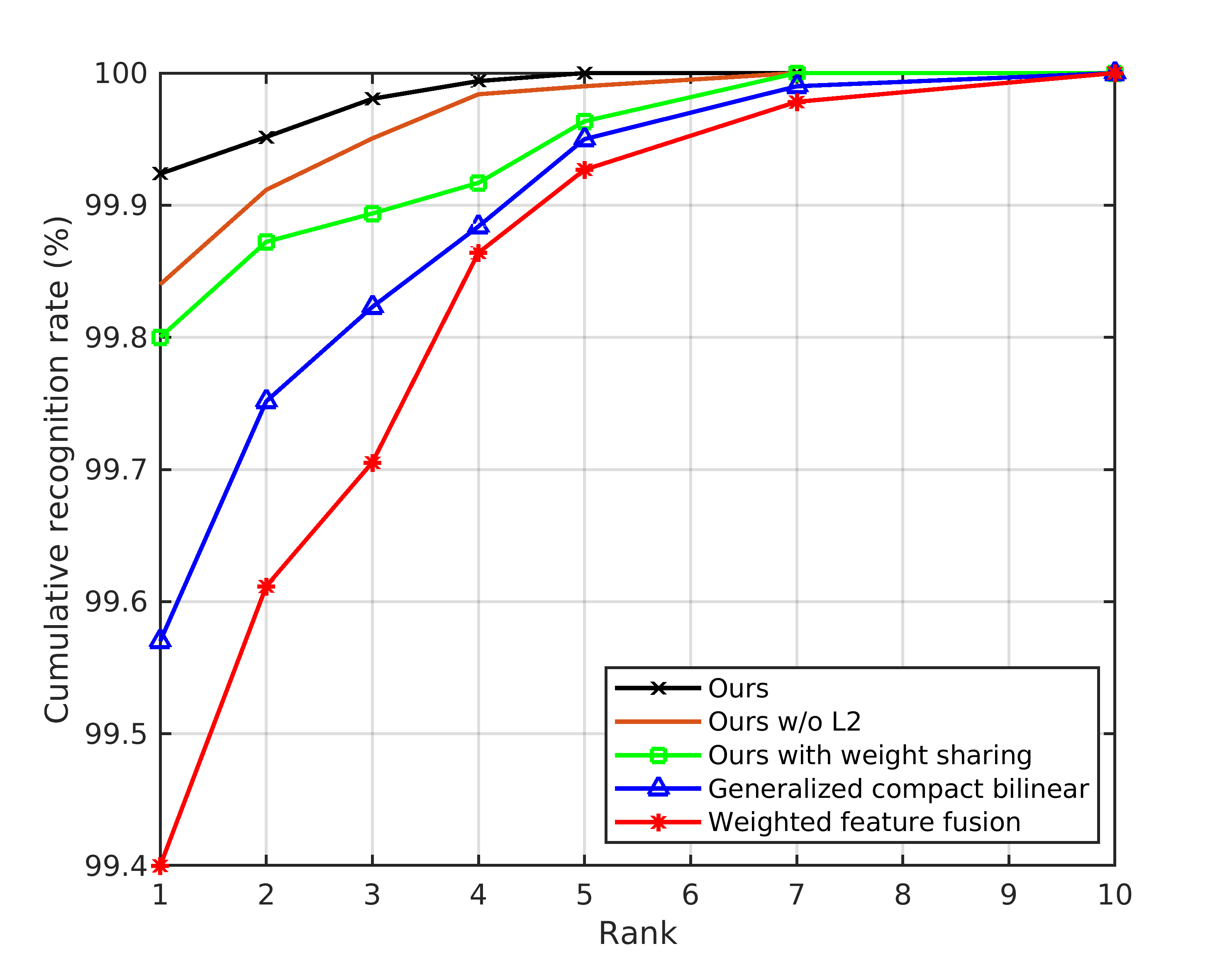}
\end{center}
\caption{The CMC curves for identification performance on the first multimodal dataset.}
\label{fig:dataset1_1}
\end{figure}

Table~\ref{table:results_dataset1} and Fig.~\ref{fig:dataset1_1} present the results for this dataset. Here, CNN-Major and CNN-Sum represent the rank-level and score-level fusion of the outputs of each modality network when quality scores for the second fusion block are not considered. As presented in Table~\ref{table:results_dataset1}, the proposed framework outperforms these two frameworks by more than $30\%$ for TAR at FAR $=10^{-4}$. We also observe that the effect of weight-sharing between $\mathrm{qNet}_k^{\mathrm{b}}$ networks, which decreases the number of parameters in the inter-modality quality score estimation networks by $66\%$ and results in $0.08$ drop of the performance in terms of EER. Fig.~\ref{fig:dataset1_1} presents the CMC curve for the proposed framework in comparison with mentioned frameworks. As presented in this figure, the proposed framework consistently outperforms the baseline frameworks. 

For this dataset, since one sample per modality is considered, the first quality-aware fusion block acts as a feature extraction framework, feeding the features to the second fusion block to estimate the inter-modality-quality of each modality. In our experiments, we observed that for this dataset the expectation of inter-modality quality score, as defined in Equation~\ref{quality_actual}, for the left iris, index, and thumb are 0.44, 0.32, and 0.24, respectively. This observation is aligned with our expectation since the unimodal performance of these three modalities, which can represent the overall quality of these modalities, follows the same sequence. Since the considered modalities are independent, we expect that the alignment of the embedding space representations of different modalities during the training should only affect the identification and not the verification performance. These expectations are consistent with the performance observed in Table~\ref{table:results_dataset1}.

\begin{figure}[t]
\begin{center}
\includegraphics[width=.99\linewidth]{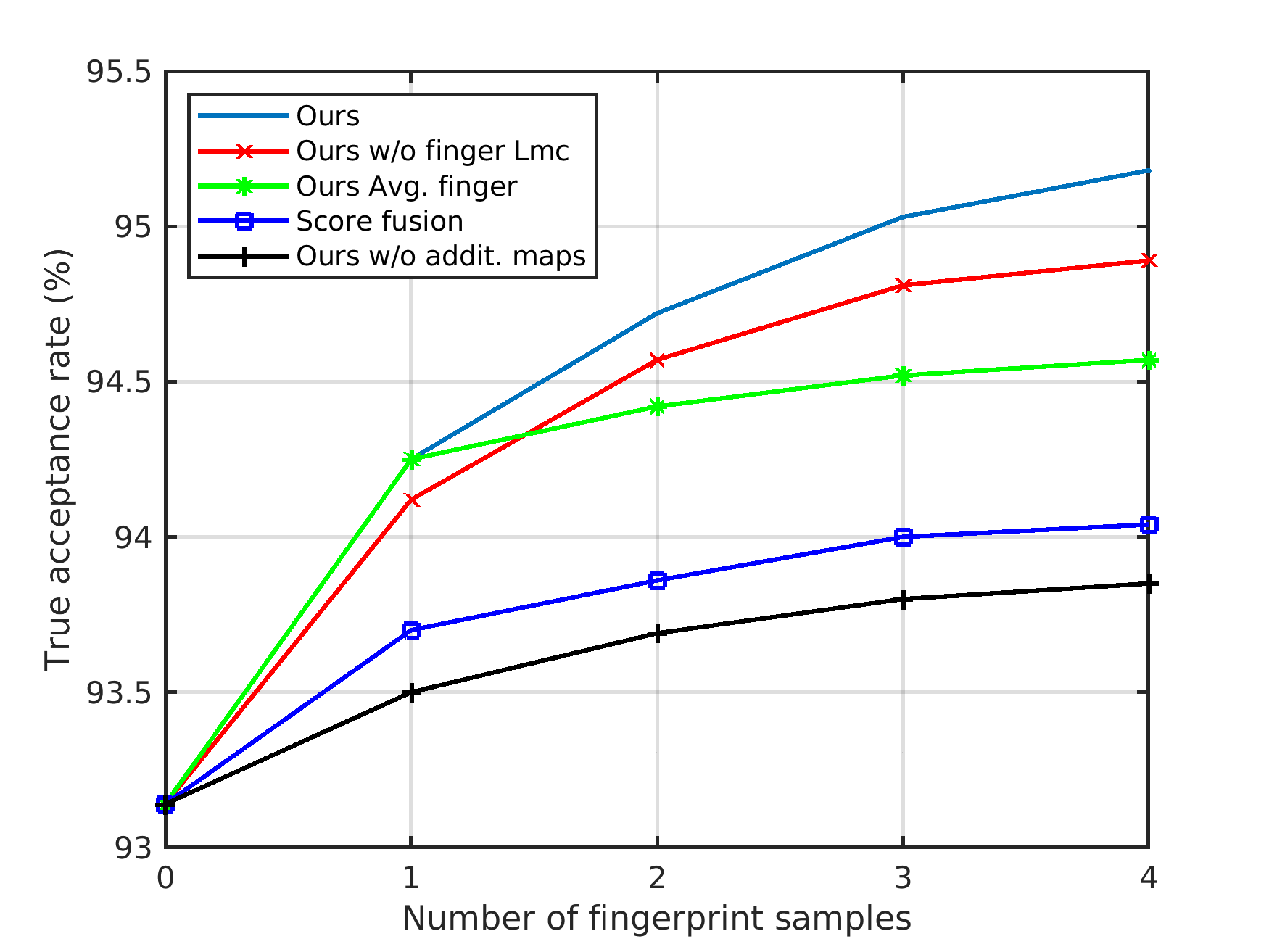}
\end{center}
\caption{The verification TAR at FAR $=10^{-3}$ for the second multimodal dataset when the number of latent fingerprint samples per subject increases.}
\label{fig:dataset2_1}
\end{figure}

\begin{figure}[t]
\begin{center}
\includegraphics[width=.95\linewidth]{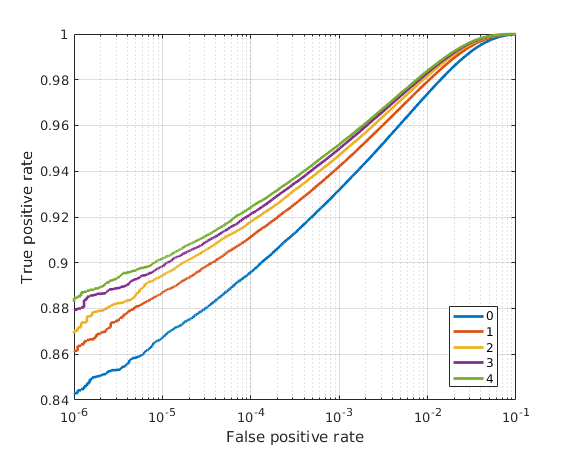}
\end{center}
\caption{The verification performance for the second multimodal dataset when the number of latent fingerprint samples per subject varies.}
\label{fig:dataset2_2}
\end{figure}

{\bf Second multimodal dataset:} This dataset consists of face samples from IJB-A~\cite{klare2015pushing} and latent fingerprints from IIIT-Delhi MOLF-Latent (D4)~\cite{sankaran2015multisensor}. IJB-A is a challenging face recognition dataset consisting of unconstrained images. This dataset contains 500 individuals, 5,397 images, and 20,412 video frames split from 2,042 videos. These images are captured with extreme pose, illumination, and expression conditions. The testing protocol for the dataset consists of 10 folds, where each fold is represented by a different random collection of 333 subjects for training and 167 for testing. IIIT-Delhi MOLF-Latent (D4) contains 4,400 fingerprint samples from 1,000 classes (10 fingers of 100 individuals). The latent fingerprints are captured using a black powder dusting process. In this experimental setup, we consider two modalities, where we assign 500 fingerprint classes with multiple samples to the IJB-A subjects. The network is fine-tuned on multimodal sample sets consisting of the remaining 500 fingerprint classes and 500 classes of  VGGFace2. As presented in Table~\ref{table:dataset2_table}, for our multimodal experiments, we follow the setup presented for the IJB-A dataset. In addition, Fig.~\ref{fig:dataset2_1} and Fig.~\ref{fig:dataset2_2} study the effect of increasing the number of fingerprint samples per subject. 

We compare the performance of the proposed framework with the same framework when the network compactness loss, $L_{mc}$, is not considered for the fingerprint modality. We observe that the AUC performance gap widens from $0.13$ to $0.29$, when the number of fingerprint samples increases from one to four. We believe this improvement is due to the more reliable fingerprint representation in the embedding space when four fingerprints are considered. We also observe that the performance when the feature vectors representing the fingerprints are averaged, i.e., the same quality score is considered for all the fingerprint samples, drops by 0.61 when considering four fingerprint samples. We study the effect of the auxiliary ridge maps when we compare the performance with the framework in which no additional map is concatenated to the original map. We observe that the variation of our framework with these additional maps in which equal quality scores are assigned to the fingerprints outperforms the previously mentioned framework with a margin of $0.72$ for four fingerprints. In addition, we compare the performance of the proposed framework with the score-based fusion of VeriFinger and $\mathrm{qNet}_\mathrm{Face}^{\mathrm{a}}$. As presented in  Table~\ref{table:dataset2_table}, inter-modality quality estimation of the proposed framework can improve the performance of the ${\mathrm{qNet}_\mathrm{Face}^\mathrm{a}}$ when combined with ${\mathrm{qNet}_\mathrm{Fing}^\mathrm{a}}$ compared to its combination with VeriFinger. 

To compare the performance of the proposed method with other fusion methods, we average the feature vectors representing each fingerprint sample and consider the score-level fusion of the fingerprint and face modalities. As presented in Fig.~\ref{fig:dataset2_1}, the proposed framework consistently outperforms the score-level fusion of the modalities. Table~\ref{table:dataset2_table} studies the effect of the inter-modality-quality score estimation by comparing the performance of the proposed framework with score-level fusion of  and $\mathrm{qNet}_\mathrm{Face}^{\mathrm{a}}$, and weighted feature fusion~\cite{soleymani2018multi} and generalized compact bilinear pooling~\cite{soleymani2018generalized} of the outputs of ${\mathrm{qNet}_k^\mathrm{a}}$ networks. As presented in this table, the conventional aggregation of the feature vectors, when one of the modalities is significantly more informative than the other modality and samples in the dataset are of varying quality, does not prove to be beneficial. On the other hand, the estimation of the inter-modality-quality scores for samples in the multimodal sample set can provide better recognition performance. As presented in this table, our proposed framework outperforms these fusion methods. However, when we study the rate of the recognition improvement by adding fingerprint samples, we observe that the gap in this rate shrinks when the number of fingerprint samples increases. For instance, adding the first fingerprint sample results in a TAR at a FAR of $10^{-3}$ improvement of $0.81$ and $0.34$ for the proposed framework and weighted feature fusion, respectively. However, the improvement, when adding the fourth sample, is equal to $0.15$ and $0.11$, respectively.

For this dataset which consists of face image sets and latent fingerprint samples of low quality, the expectations for inter-modality quality scores, when one latent fingerprint is considered, are 0.72 and 0.28 respectively. The expectation for inter-modality-quality scores of the fingerprint modality scores increases from $0.28$ to $0.33$ when the number of fingerprints per multimodal sample set increases from one to four. These inter-modality quality scores, which can represent both the importance of a modality in the joint multimodal framework as well as the overall quality of the samples, are aligned with our expectation about inter-modality-quality scores.

{\bf Third multimodal dataset:} We construct this multimodal dataset using three datasets. YouTube Face dataset (YTF)~\cite{wolf2011face} is designed for unconstrained face verification in videos. It contains 3,425 videos of 1,595 different people, and the video lengths vary from 48 to 6,070 frames with an average length of 181.3 frames. In this dataset, ten folds of 500 video pairs are available. CASIA Iris V4-Distance dataset is a subset of database~\cite{CasiaIrisV4} and contains 2,446 instances from 142 different subjects. IIIT-Delhi Latent fingerprint (D4) dataset~\cite{sankaran2011matching} consists of 1046 latent fingerprint samples pertaining to 15 subjects with all 10 fingerprints, thus the dataset has 150 classes. The latent fingerprints are captured under semi-controlled environment the black powder dusting process. The dataset is prepared in multiple sessions with variations in background, and captures the effect of dryness, wetness, and moisture. This provides sample variation in the quality, noise, and information content of latent fingerprint samples. The samples of lifted latent fingerprints are digitized using a Canon EOS 500D camera. For our third dataset, we consider three modalities and 142 classes. Here, we select 142 random classes from YouTube Face, assign iris classes and randomly selected fingerprint classes to them, and follow the protocol described in YouTube Face for our verification setup. The network is fine-tuned using the VGGFACE2, the right iris samples from CASIA Iris V4-Distance, and the remaining subjects from IIIT-Delhi Latent (D4).

\begin{figure}[t]
\begin{center}
\includegraphics[width=.95\linewidth]{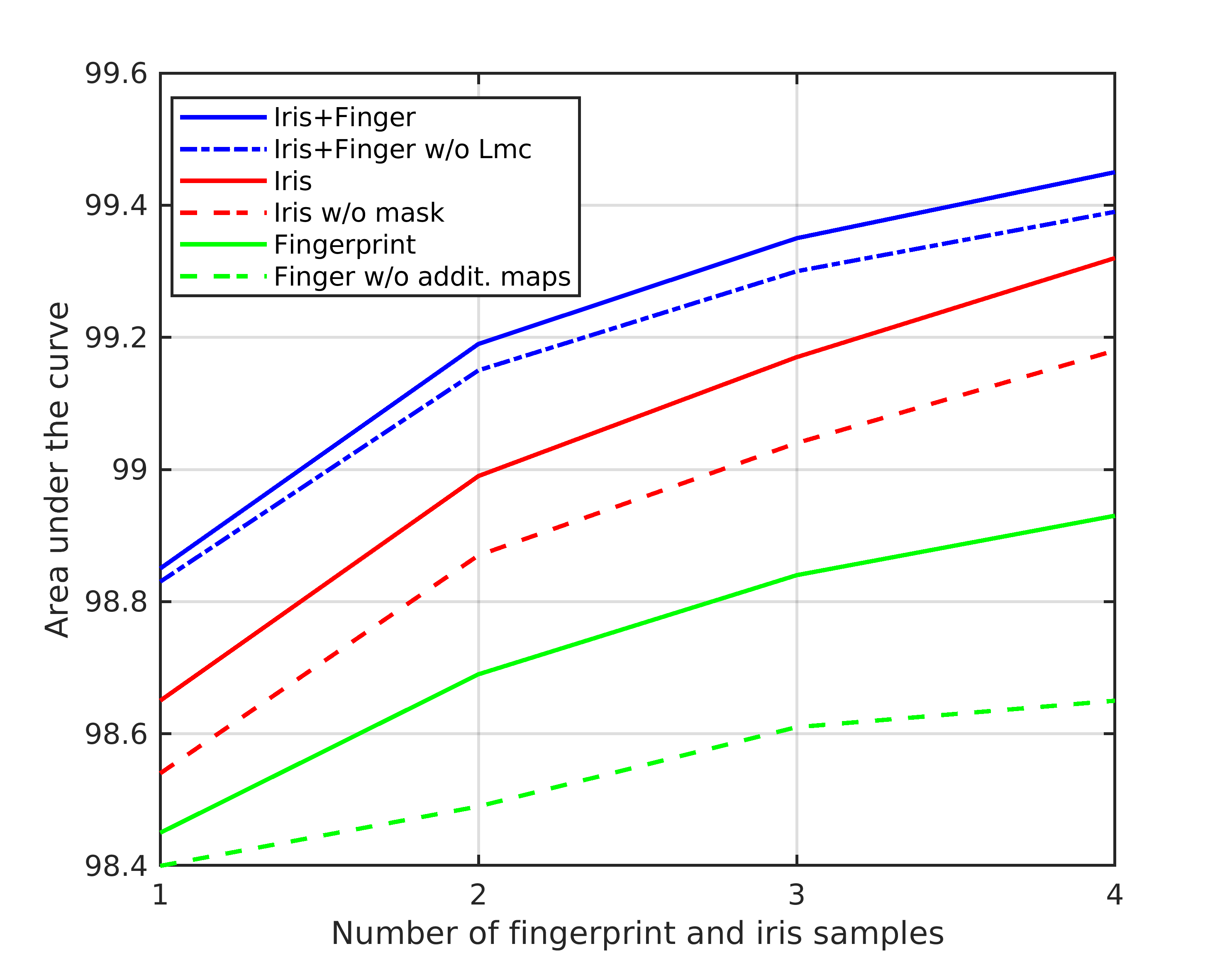}
\end{center}
\caption{The verification AUC for the third multimodal dataset when the number of iris and latent fingerprint samples per subject increases.}
\label{fig:dataset3_1}
\end{figure}

Table~\ref{table:results_dataset3_all} and Fig.~\ref{fig:dataset3_1} present the verification performance for this dataset. In Table~\ref{table:results_dataset3}, we study the impact of adding the latent fingerprint and iris samples to the YouTube Face dataset classes. As expected, adding iris samples improves the recognition performance better than adding the latent fingerprint samples. In addition, the performance improves drastically after adding the first few fingerprint and iris samples. However, this improvement lessens when adding several samples. Furthermore, Fig.~\ref{fig:dataset3_1} presents results related to the effect of different modifications in the multimodal framework. In this figure, the blue curves present the results when up to four pairs of iris and latent fingerprint samples are added to each subject. Similarly, red and green curves represent the performance when adding only iris or fingerprint samples to each class, respectively. The dashed blue curve represents the performance when $L_{mc}$ is not considered for iris and fingerprint networks. As presented in this figure, the iris modality outperforms the latent fingerprint modality by a wide margin. In addition, the gap between the improvement resulted by using iris samples and latent fingerprint samples widens as the number of considered samples increases. On the other hand, the performance of the framework in which only iris samples are added to YouTube Face dataset classes approaches the performance of framework using all three modalities as the number of iris samples increases.  

For this multimodal dataset, we observe that the inter-modality-quality score for face, iris and fingerprint modalities, when one fingerprint and iris samples are included in the multimodal sample set, are 0.51, 0.30, and 0.19, respectively. However, by increasing the number of iris and fingerprint samples from one to four, their corresponding scores improve to $0.37$ and $0.24$, respectively. As studied in~\cite{deng2019arcface}, the combination of inter-class loss functions cannot improve the angular loss defined in Equation~\ref{angualrloss} for face recognition. However, we observe that the multimodal network compactness loss, $L_{mc}$, although not very efficient for the unimodal networks, can improve the multimodal performance. We believe this is due to the over-parametrization of the multimodal network compared to the unimodal networks. In addition, we compare the performance of the proposed framework with the score-level fusion of VeriFinger and OSIRIS in Table~\ref{table:results_dataset3_verifinger}. As presented in this table, the proposed framework outperforms the performance of this score-level fusion with a wide margin when considering a fewer number of samples. However, when the number of samples is increased, the performance of the score-level fusion becomes closer to the proposed framework. 

\begin{table}[t]
\caption[Table caption text]{The verification AUC for the third multimodal dataset when the number of latent fingerprint and iris samples per subject vary.}
\begin{subtable}[h]{0.5\textwidth}
\begin{center}
\begin{tabular}{l@{\hskip .05in}cc@{\hskip .05in}|c@{\hskip .05in}|c@{\hskip .05in}|c@{\hskip .05in}|c}
\rowcolor{white}\multicolumn{2}{c}{\cellcolor{white}{}}{ } &\multicolumn{5}{c}{Iris}\\ 
\rowcolor{black!20}\multicolumn{2}{c}{\cellcolor{white}{}}{ }& 0&  1 & 2 &3&4 \\ 
{\cellcolor{white}{}}&{\cellcolor{black!20}{0}}&98.12&98.65&98.99&99.17&99.32\\ 
{\cellcolor{white}{}}          &{\cellcolor{black!20}{1}}&98.45& 98.85& 99.10& 99.25& 99.39\\ 
{\cellcolor{white}{}}                              &{\cellcolor{black!20}{2}}&98.69& 99.01& 99.19& 99.31& 99.42\\ 
{\cellcolor{white}{}}           &{\cellcolor{black!20}{3}}&98.84& 99.12& 99.25& 99.35& 99.44\\ 
\multirow{-5}{*}{\rotatebox[origin=c]{90}{Fingerprint}} {\cellcolor{white}{}} &{\cellcolor{black!20}{4}}&98.93& 99.17& 99.28& 99.38& 99.45\\ 
\end{tabular}
\end{center}
\caption[Table caption text]{Ours}
\label{table:results_dataset3}
\end{subtable}
\begin{subtable}[h]{0.5\textwidth}
\begin{center}
\begin{tabular}{l@{\hskip .05in}cc@{\hskip .05in}|c@{\hskip .05in}|c@{\hskip .05in}|c@{\hskip .05in}|c}
\rowcolor{white}\multicolumn{2}{c}{\cellcolor{white}{}}{ } &\multicolumn{5}{c}{Iris}\\ 
\rowcolor{black!20}\multicolumn{2}{c}{\cellcolor{white}{}}{ }& 0&  1 & 2 &3&4 \\ 
{\cellcolor{white}{}}&{\cellcolor{black!20}{0}}&98.12&98.23&98.44&98.53&98.61\\ 
{\cellcolor{white}{}}          &{\cellcolor{black!20}{1}}&98.24& 98.42& 98.53& 98.61& 98.75\\ 
{\cellcolor{white}{}}                              &{\cellcolor{black!20}{2}}&98.43& 98.61& 98.82& 98.72& 98.94\\ 
{\cellcolor{white}{}}           &{\cellcolor{black!20}{3}}&98.67& 98.82& 98.94& 99.03& 99.12\\ 
\multirow{-5}{*}{\rotatebox[origin=c]{90}{Fingerprint}} {\cellcolor{white}{}} &{\cellcolor{black!20}{4}}&98.72& 98.93& 99.13& 99.26& 99.36\\ 
\end{tabular}
\end{center}
\caption[Table caption text]{${\mathrm{qNet}_\mathrm{Face}^\mathrm{a}}$+VeriFinger+OSIRIS-Sum.}
\label{table:results_dataset3_verifinger}
\end{subtable}
\label{table:results_dataset3_all}
\end{table}

{\bf Chimeric pairing and statistical analysis:} Due to the lack of multimodal datasets for evaluating the current work, we manually constructed two chimeric multimodal sets. Multimodal samples generated randomly in our experiments can possess varying recognition potentials which consequently affects the matching performance and comparisons. To provide a better performance assessment for chimeric datasets, we expanded our experiments by evaluating the standard deviation of the multimodal recognition performance over five different sets of virtual subjects from the unimodal datasets. The standard deviation of the performance for the second and the third datasets, when considering four samples from each modality, is observed to be less than 0.05\% and 0.02\%, respectively. This validates the effectiveness of random pairing for constructing multimodal datasets using independent single modalities, and advocates that the same criterion can be used to alleviate the scarcity of real-world multimodal datasets.
It is also worth mentioning that, although the improvement in the performance resulted from multimodal settings compared to the unimodal settings, may be considered as small, these improvements can be beneficial when large-scale applications such as passport control are considered~\cite{kemelmacher2016megaface,wolf2011face,huang2008labeled,sultana2017social}.

\subsection{Ablation study}
Here, we study the feature extraction performance of the first quality-aware fusion block when a single sample is fed to the network. On the other hand, the aggregation capability and quality estimation performance of this block is studied when we feed multiple samples to the unimodal network.

{\bf Single-sample and single-modality:} For the IJB-A~\cite{klare2015pushing} and YouTube Face~\cite{wolf2011face} datasets, we follow the protocol presented in the corresponding papers. For the identification setup for iris datasets contained in BIOMDATA~\cite{crihalmeanu2007protocol} and CASIA-Distance~\cite{CasiaIrisV4}, we consider the protocol presented in~\cite{haghighat2016discriminant}, while for the verification setup we follow~\cite{zhao2017towards}. Here, in the identification setup, for each subject, four randomly selected samples are considered as the gallery and the remaining samples are considered as the probes. For fingerprint datasets present in BIOMDATA, IIIT-Delhi MOLF-Latent (D4)~\cite{sankaran2015multisensor}, and the IIIT-Delhi Latent fingerprint dataset~\cite{sankaran2011matching}, we follow~\cite{haghighat2016discriminant},~\cite{sankaran2015multisensor}, and~\cite{sankaran2011matching}, respectively. It should be noted that, for experiments performed on the IIIT-Delhi MOLF-Latent (D4) dataset, we consider the latent-to-sensor framework, while for the IIIT-Delhi Latent fingerprint dataset, latent-to-latent recognition is considered. 

\begin{table}[t]
\caption[Table caption text]{The unimodal performance of $\mathrm{qNet}^{\mathrm{a}}_\mathrm{Iris}$ on single-samples. For Sun {\it et al.} and Zhao {\it et al.}, results are reported from~\cite{zhao2017towards}.}
\begin{center}
\begin{tabular}{l@{\hskip .05in}c@{\hskip .05in}c@{\hskip .05in}c@{\hskip .05in}|c@{\hskip .05in}c@{\hskip .05in}c@{\hskip .05in}c@{\hskip .05in}c@{\hskip .05in}c@{\hskip .05in}}
\toprule 
&\multicolumn{3}{c}{BIOMDATA Left}&\multicolumn{3}{c}{CASIA-Dist. Left}\\
Method &  @$10^{-3}$ & Rank-1 &EER &  @$10^{-3}$ & Rank-1 & EER \\ \hline
OSIRIS~\cite{othman2016osiris}            &86.30&95.17&4.43&         80.07&88.68&6.39\\
Sun {\it et al.}~\cite{sun2008ordinal}    &90.11&--&5.19&         83.07&--&7.89\\
Zhao {\it et al.}~\cite{zhao2017towards}  &94.30&--&2.63&         84.10&--&5.50\\
{Ours} w/o mask                           &92.84&99.38&3.14&         83.51&94.52&6.47\\
{Ours}                                    &{\bf 95.48}&{\bf 99.51}&{\bf 2.15}&         {\bf 86.12}&{\bf 96.14}&{\bf5.09}\\
\hline
\end{tabular}
\end{center}
\label{table:results_singleM_singleS-FaceIris}
\end{table}

\begin{table*}[t]
\caption[Table caption text]{The unimodal performance of ${\mathrm{qNet}_\mathrm{Fing}^\mathrm{a}}$ for single-samples.}
\begin{center}
\begin{tabular}{l@{\hskip .05in}c@{\hskip .05in}c@{\hskip .05in}c@{\hskip .05in}||c@{\hskip .05in}c@{\hskip .05in}c@{\hskip .05in}||c@{\hskip .05in}c@{\hskip .05in}c@{\hskip .05in}||c@{\hskip .05in}c@{\hskip .05in}c@{\hskip .05in}c@{\hskip .05in}}
\toprule 
&\multicolumn{3}{c}{BIOMDATA L Thumb}&\multicolumn{3}{c}{BIOMDATA L Index}&\multicolumn{3}{c}{MOLF-Latant}&\multicolumn{4}{c}{IIIT-Latant}\\
\toprule 
Method &  @$10^{-3}$ & EER & Rank-1 &  @$10^{-3}$ & EER & Rank-1&@$10^{-2}$ &  Rank-25 &Rank-50&  @$10^{-2}$ & Rank-1 & Rank-10 &Rank-25  \\ 
NBIS~\cite{ko2007users}            &57.49&14.45&71.23   &68.68&6.48&87.17  &12.21&5.01&8.63    &48.33&52.31&58.90&63.42\\
VeriFinger~\cite{VeriFinger}       &68.26&12.03&76.16   &76.12&6.18&90.41  &8.14&2.87&6.56    &55.19&61.02&74.00&77.44\\
Ours w/o 9 maps                    &61.23&13.16&74.78   &73.12&6.35&88.51  &5.21&4.51&7.73    &52.89&57.57&68.21&72.35\\
{Ours} w/o $L_{mc}$                &74.83&8.71&83.43    &81.84&5.87&94.87  &22.12&37.14&64.94  &60.56&69.47&81.54&86.81\\
{Ours}                             &{\bf76.16}&{\bf 7.52}&{\bf 84.71    }&{\bf 87.63}&{\bf 4.66}&{\bf 95.12  }&{\bf 25.14}&{\bf 40.41}&{\bf 67.11  }&{\bf 65.53}&{\bf 73.81}&{\bf 85.21}&{\bf89.74}\\
\hline
\end{tabular}
\end{center}
\label{table:results_singleM_singleS-Fing}
\end{table*}

\begin{table*}[t]
\caption[Table caption text]{Identification performance of ${\mathrm{qNet}_\mathrm{Fing}^\mathrm{a}}$ for multi-sample setup consisting of one to four samples.}
\begin{center}
\begin{tabular}{l@{\hskip .055in}c@{\hskip .055in}c@{\hskip .055in}c@{\hskip .055in}c@{\hskip .055in}||c@{\hskip .055in}c@{\hskip .055in}c@{\hskip .055in}c@{\hskip .055in}||c@{\hskip .055in}c@{\hskip .055in}c@{\hskip .055in}c@{\hskip .055in}||c@{\hskip .055in}c@{\hskip .055in}c@{\hskip .055in}c@{\hskip .055in}}
\toprule 
&\multicolumn{4}{c}{BIOMDATA L Thumb}&\multicolumn{4}{c}{BIOMDATA L Index}&\multicolumn{4}{c}{MOLF-Latant}&\multicolumn{4}{c}{IIIT-Latant}\\
&\multicolumn{4}{c}{Rank-1}&\multicolumn{4}{c}{Rank-1}&\multicolumn{4}{c}{Rank-25}&\multicolumn{4}{c}{Rank-1}\\
\toprule 
Method & 1& 2 & 3 & 4 & 1& 2 & 3 & 4&1&2 & 3 & 4& 1& 2 & 3 & 4\\ 
VeriFinger-Major    &76.16&76.74&82.45&84.23&     90.41&91.22&94.92&95.58&    2.87&3.54&9.10 &13.41&    61.02&62.81&62.84&76.20\\
VeriFinger-Sum      &76.16&80.10&83.87&85.41&     90.41&93.98&95.15&95.89&    2.87&5.49&10.83&14.45&    61.02&68.82&74.43&78.56\\

Ours-w/o 9 maps-Sum &74.78&76.45&77.34&78.12&     88.51&90.38&91.79&92.24&    4.51&5.81&6.76&7.13&      57.57&61.79&63.64&65.42\\
Ours-w/o 9 maps     &74.78&78.51&81.64&83.34&     88.51&91.56&93.58&94.19&    4.51&6.03&8.42&9.05&      57.57&64.80&69.28&63.82\\

Ours-Major          &84.71&85.12&91.53&93.85&     95.12&95.47&97.63&98.51&   40.41&40.92&56.52&60.33&   73.81&73.91&80.95&82.49\\
Ours-Sum            &84.71&87.74&91.63&93.59&     95.12&97.96&98.01&98.94&   40.41&50.48&56.33&60.07&   73.81&77.23&80.18&82.41\\
Ours-Avg.           &84.71&88.57&92.46&94.12&     95.12&98.17&98.87&99.22&   40.41&51.39&57.27&62.76&   73.81&77.40&81.42&83.83\\
Ours                &{\bf84.71}&{\bf89.41}&{\bf93.74}&{\bf95.64}&{\bf     95.12}&{\bf98.53}&{\bf99.23}&{\bf99.70}&{\bf   40.41}&{\bf52.34}&{\bf58.79}&{\bf64.88}&{\bf   73.81}&{\bf78.78}&{\bf82.64}&{\bf85.12}\\
\hline
\end{tabular}
\end{center}
\label{table:results_singleM_MultiS_2}
\end{table*}

\begin{table*}[t]
\caption[Table caption text]{Recognition performance of ${\mathrm{qNet}_\mathrm{Iris}^\mathrm{a}}$ for multi-sample setup consisting of one to four samples.}
\begin{center}
\begin{tabular}{l@{\hskip .075in}c@{\hskip .075in}c@{\hskip .075in}c@{\hskip .075in}c@{\hskip .075in}||c@{\hskip .075in}c@{\hskip .075in}c@{\hskip .075in}c@{\hskip .075in}||c@{\hskip .075in}c@{\hskip .075in}c@{\hskip .075in}c@{\hskip .075in}||c@{\hskip .075in}c@{\hskip .075in}c@{\hskip .075in}c@{\hskip .075in}}
\toprule 
&\multicolumn{8}{c}{BIOMDATA Left}&\multicolumn{8}{c}{CASIA-Dist. Left}\\
&\multicolumn{4}{c}{@$10^{-3}$}&\multicolumn{4}{c}{Rank-1}&\multicolumn{4}{c}{@$10^{-3}$}&\multicolumn{4}{c}{Rank-1}\\
\toprule 
Method & 1& 2 & 3 & 4 & 1& 2 & 3 & 4&1&2 & 3 & 4& 1& 2 & 3 & 4\\ 
OSIRIS-Major        &86.30&86.41&88.41&90.17&    95.17&95.21&95.83&95.97&   80.07&80.23&85.68&86.35&   88.68&88.70&89.01&89.12\\
OSIRIS-Sum          &86.30&89.10&90.92&91.53&    95.17&95.94&96.43&96.78&   80.07&84.49&86.83&88.45&   88.68&88.88&89.15&89.35\\
Ours-Major          &95.48&97.02&97.12&97.34&    99.51&99.54&99.55&99.55&   86.12&86.22&86.32&86.38&   96.14&96.63&96.93&97.13\\
Ours-Sum            &95.48&97.74&97.63&97.59&    99.51&99.55&99.07&99.59&   86.12&88.33&89.41&90.45&   96.14&96.75&96.82&97.25\\
Ours-Avg.           &95.48&97.49&98.17&98.55&    99.51&99.60&99.63&99.67&   86.12&88.87&91.24&93.85&   96.14&96.87&97.57&97.92\\
Ours                &{\bf95.48}&{\bf98.67}&{\bf99.75}&{\bf99.82}&{\bf    99.51}&{\bf99.62}&{\bf99.67}&{\bf99.71}&{\bf   86.12}&{\bf89.58}&{\bf92.81}&{\bf94.32}&{\bf   96.14}&{\bf97.35}&{\bf98.51}&{\bf98.89}\\
\hline
\end{tabular}
\end{center}
\label{table:results_singleM_MultiS_3}
\end{table*}

Tables~\ref{table:results_singleM_singleS-FaceIris} and ~\ref{table:results_singleM_singleS-Fing} present the unimodal results for the single-sample setup on iris and fingerprint modalities. The performance of the ${\mathrm{qNet}_\mathrm{Iris}^\mathrm{a}}$ is compared to OSIRIS~\cite{othman2016osiris}, Sun {\it et al.}~\cite{sun2008ordinal}, and Zhao {\it et al.}~\cite{zhao2017towards}. We also compare the performance of the proposed framework with the same framework when not concatenating the mask images to the iris images. As presented in Table~\ref{table:results_singleM_singleS-FaceIris}, concatenating the mask image with the iris image can improve the recognition performance, which outperforms the state-of-the-art framework~\cite{zhao2017towards} by $0.48$ and $0.41$ in terms of EER on BIOMDATA left and CASIA-Distance left, respectively.

\begin{table*}[t]
\caption[Table caption text]{The performance of $\mathrm{qNet}^{\mathrm{a}}_\mathrm{Face}$ for multi-sample face recognition.}
\begin{center}
\begin{tabular}{lcc|cc||cc}
\toprule 
&\multicolumn{4}{c}{IJB-A}&\multicolumn{2}{c}{YTF}\\
\toprule 
&\multicolumn{2}{c}{Verification}&\multicolumn{2}{c}{Identification}&\multicolumn{2}{c}{Verification}\\
Method & @$10^{-2}$ & @$10^{-3}$ & Rank-1 &Rank-5& @$10^{-3}$ &  AUC\\ 
\hline
DR-GAN~\cite{tran2017disentangled}&$77.4\pm2.7$&$53.9\pm4.3$&$85.5\pm1.5$&$94.7\pm1.1$&--&--\\
Triplet Similarity~\cite{sankaranarayanan2016triplet}&$79.0\pm3.0$&$59.0\pm5.0$&$88.0\pm1.5$&$95.0\pm0.7$&--&--\\
Template Adaptation~\cite{crosswhite2018template}&$93.9\pm1.3$&$83.6\pm2.7$&$92.8\pm1.0$&$97.7\pm0.4$&--&--\\
NAN~\cite{yang2017neural}&$93.3\pm0.9$&$86.0\pm1.2$&$95.4\pm0.7$&$97.8\pm0.4$&$95.72$&$98.8$\\
SphereFace~\cite{liu2017sphereface}&$92.3\pm1.6$&$88.4\pm4.2$&$93.2\pm1.3$&$96.5\pm1.1$&$95.0$&--\\
DeepFace~\cite{taigman2014deepface}&--&--&--&--&91.4&96.3\\
CosFace~\cite{wang2018cosface}&--&--&--&--&97.6&--\\
ArcFace~\cite{deng2019arcface}&--&--&--&--&98.02&--\\
PRN~\cite{kang2018pairwise}&$96.5\pm0.4$&$91.9\pm1.3$&$98.2\pm0.4$&{\bf 99.2} $\pm$ {\bf 0.2}&95.8&--\\
UniformFace~\cite{duan2019uniformface}&$96.9\pm0.8$&$92.3\pm1.7$&$97.9\pm0.5$&$98.8\pm0.2$&$97.7$&--\\
\hline 
Ours-Avg.         &$92.5\pm0.8$&$82.7\pm2.3$    &$97.9\pm0.5$&$98.8\pm0.2$&     $97.51$&$99.0$\\
Ours w/o $L_{mc}$&$97.0\pm0.7$&$92.5\pm1.9$    &$98.0\pm0.5$&$98.8\pm0.3$&      $98.06$&$99.0$\\
Ours             &{\bf 97.3} $\pm$ {\bf0.7}&{\bf 93.1} $\pm$ {\bf1.7} &{\bf 98.4} $\pm$ {\bf 0.4}&{\bf 99.2} $\pm$ {\bf 0.2}&{\bf      98.12}&{\bf 99.1}\\
\hline
\end{tabular}
\end{center}
\label{table:results_singleM_MultiS_1}
\end{table*}

Table~\ref{table:results_singleM_singleS-Fing} presents the performance of ${\mathrm{qNet}_\mathrm{Fing}^\mathrm{a}}$ compared to NBIS~\cite{ko2007users}, VeriFinger~\cite{VeriFinger}, and ${\mathrm{qNet}_\mathrm{Fing}^\mathrm{a}}$ fed with only the original fingerprint image processed with Gabor filters with locally estimated angles and not concatenated with the remaining nine maps explained in the {\bf Data representation} section. As presented in this table, the additional ridge maps are mostly beneficial for latent fingerprints since these maps are constructed using constant directions for the whole fingerprint image. These auxiliary maps are considered according to the fact that ridge orientation estimation for latent fingerprints is not accurate due to the photometric and geometric distortions, and therefore, using that orientation map for enhancement reduces the reliability of ridge information. However, by considering a constant global direction for each auxiliary map, there is a chance that the ridge information from unreliable regions can be enhanced in at least one of the additional maps and the deep model can exploit these maps to construct better representations.

{\bf Multi-sample and single-modality:} Tables~\ref{table:results_singleM_MultiS_1},~\ref{table:results_singleM_MultiS_2}, and~\ref{table:results_singleM_MultiS_3} present the results for the unimodal multi-sample setup for face, fingerprint, and iris modalities, respectively. The unimodal networks are trained with uniform angular loss as defined by Equation~\ref{uniformsoftmax} and network compactness loss. In addition to comparison with other frameworks, in these tables, we study the effect of these loss functions on the unimodal performance. As presented in Table~\ref{table:results_singleM_MultiS_1}, the performance of the ${\mathrm{qNet}_\mathrm{Face}^\mathrm{a}}$ is compared to frameworks such as SphereFace~\cite{liu2017sphereface} and {UniformFace}~\cite{duan2019uniformface}, and it is shown to achieve state-of-the-art performance with fewer training samples; 3M compared to 6M for UniformFace. Here, on the face recognition task, we clearly observe the effect of the network compactness loss as well as the separability loss which improve the performance of the proposed framework to $93.1\%$ and $98.12\%$ on IJB-A and YouTube Face datasets for verification at a FAR of $10^{-3}$, respectively.  

Table~\ref{table:results_singleM_MultiS_2} presents the multi-sample recognition results on four fingerprint datasets when up to four samples are considered. The performance of the proposed quality-aware framework is compared to the rank-level and score-level fusions of the single-sample representations as well as element-wise averaging of these representations. On the other hand, the fusion of the VeriFinger scores outperforms our framework when auxiliary ridge maps are not considered. However, the proposed framework outperforms VeriFinger by a large margin on the latent fingerprints. Table~\ref{table:results_singleM_MultiS_3} presents the results for BIOMDATA left and CASIA-Distance left iris datasets. As presented in this table, the proposed framework consistently outperforms different score-, rank-, and feature-level variations of the multi-sample frameworks as well as score-level and rank-level fusion of the OSIRIS verifier.

\begin{figure}[t]
\begin{center}
\includegraphics[width=.90\linewidth]{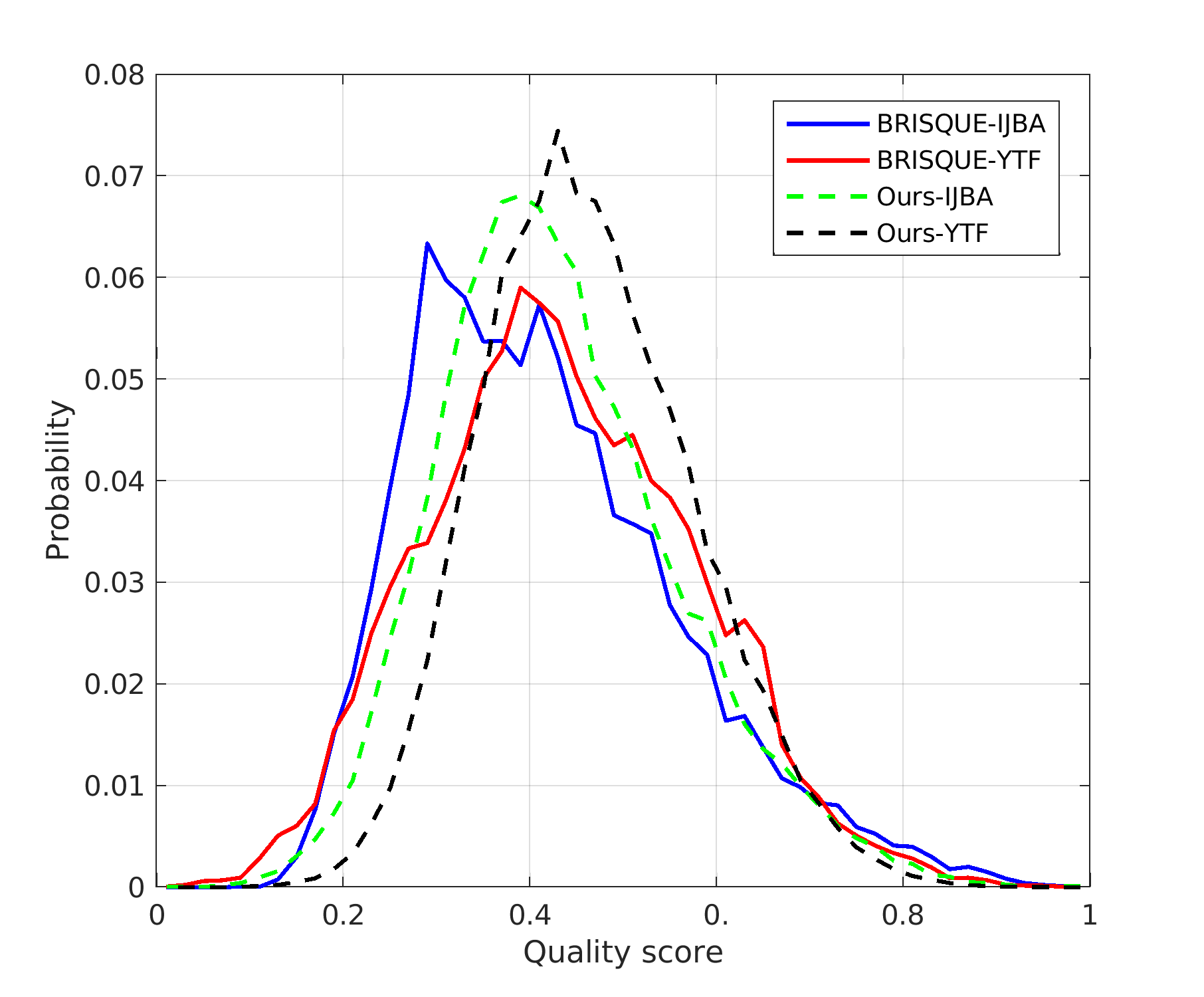}
\end{center}
\caption{The probability of estimated quality scores for the proposed framework compared to BRISQUE on IJB-A and  Youtube Face datasets.}
\label{fig:BRISQUE}
\end{figure}

\begin{figure}[t]
\begin{center}
\includegraphics[width=.95\linewidth]{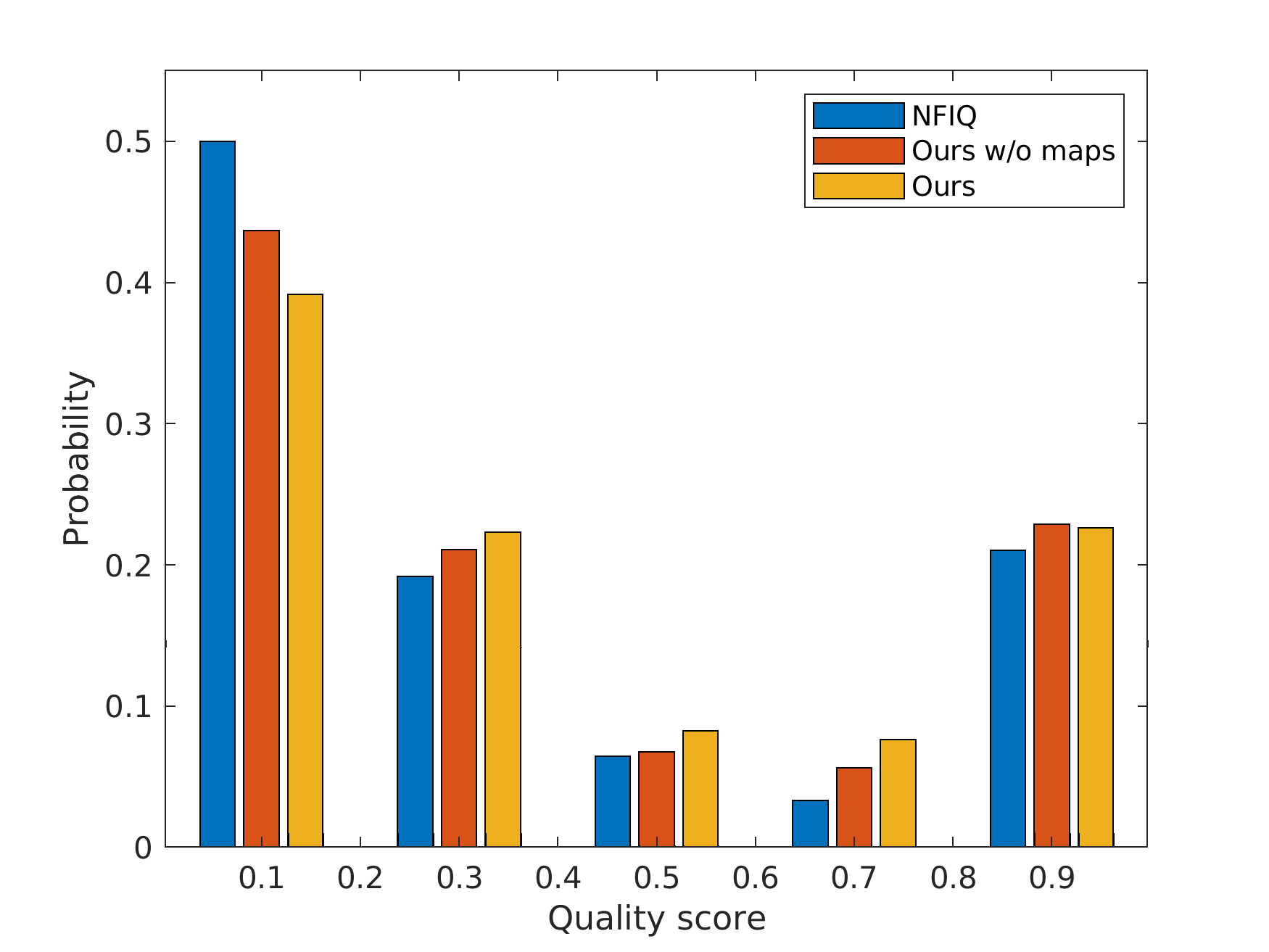}
\end{center}
\caption{The probability of {\it quantized} quality scores for the proposed framework compared to NFIQ 2.0 on fingerprint datasets.}
\label{fig:NFIQ}
\end{figure}

{\bf Quality measures: }To analyze the quality scores estimated by the proposed framework, we focus on our single-sample single-modality frameworks and the comparison of the distribution of learned scores with no-reference image quality metrics Blind/Referenceless Image Spatial Quality Evaluator (BRISQUE)~\cite{mittal2011blind} and NIST Fingerprint Image Quality (NFIQ 2.0)~\cite{bausinger2011fingerprint}. The BRISQUE score is predicted by a support vector regression model trained on a set of images with their corresponding differential mean opinion score as the target. The set of images contain original images along with their distorted versions corrupted by known distortion effects such as compression artifacts, blurring, and noise. The BRISQUE score is a scalar value typically in the range $[0,100]$. Lower values of BRISQUE score reflects better perceptual quality. NFIQ 2.0 assigns each fingerprint a score from zero (low quality) to 100 (high quality). In Figs~\ref{fig:BRISQUE} and~\ref{fig:NFIQ}, to provide a better comparison, all the quality scores are normalized to $[0,1]$, with higher values representing better quality. It is worth mentioning that in the original BRISQUE algorithm lower values represent higher qualities. Therefore, in Fig~\ref{fig:BRISQUE}, the BRISQUE scores are reversed as, {\it 1-normalized score}, to provide comparison with the scores estimated by other frameworks.     

Fig.~\ref{fig:BRISQUE} compares the distribution of the quality scores measured by the BRISQUE with our multi-sample single-modality framework on IJB-A and Youtube Face datasets. As presented in this figure, the distribution of the quality scores estimated by the proposed framework closely follow the BRISQUE scores. On the other hand, Fig.~\ref{fig:NFIQ} compares the quality scores estimated by the proposed framework with NFIQ 2.0. In this figure, all the fingerprint samples from three test multimodal datasets are considered and the scores are {\it quantized} to five levels. As presented in these figures, our estimated scores are closer to one compared to no-reference measures since our proposed networks are trained mostly on corrupted samples, while NFIQ 2.0 and BRISQUE are trained on a large number of images with various qualities including high-quality samples. Therefore, our estimated scores provide a relative quality measure for low-quality samples, while each mentioned no-reference measure computes a more global quality metric. It should be noted that our quality scores do not aim to serve as stand-alone quality measures, and the current comparisons seek to showcase the positive correlation of our sample weighting scores with actual quality scores.

\begin{figure*}[t]
\begin{center}
\includegraphics[width=.98\linewidth]{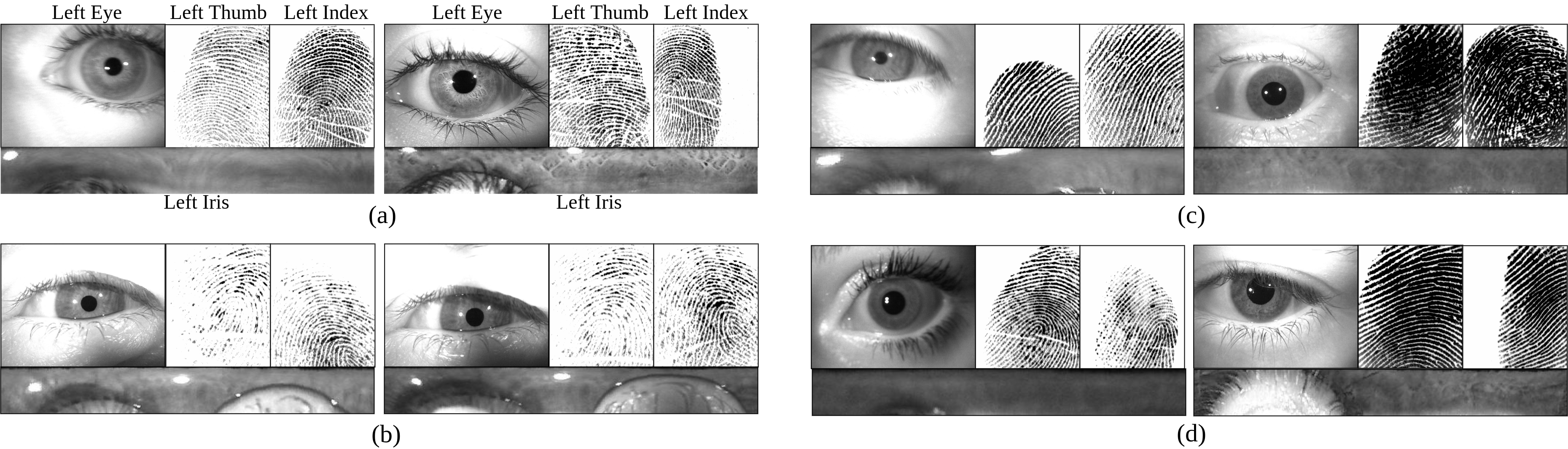}
\end{center}
\caption{Examples of failed verification for the first multimodal dataset, (a-b) false rejection at FAR = $10^{-4}$ and (c-d) false acceptance at FRR = $10^{-3}$.}
\label{fig:failed_samples}
\end{figure*}

{\bf Failed verification of sample sets: }To study the performance of the proposed framework from another perspective, we investigate the multimodal sample sets that our framework fails to correctly verify. The first  multimodal dataset which consists of BIOMDATA samples, includes images corrupted with blur, occlusion, shadows, and sensor noise. In Fig.~\ref{fig:failed_samples}, we present four pairs of sample sets, consisting of left iris, left thumb, and left index, that the framework fails to correctly verify. For better presentation, we also present the left eye, although the eye image is not used as an input to the framework. In Figs~\ref{fig:failed_samples}(a) and~\ref{fig:failed_samples}(b), we investigate positive multimodal sample pairs that are rejected by our framework for false acceptance rate (FAR) of $10^{-4}$. To further investigate the performance of the framework, we increase the FAR for these sample sets and observe that these pairs are correctly verified at FAR = $2\times 10^{-4}$ and $5\times 10^{-4}$, respectively. Then, in Figs~\ref{fig:failed_samples}(c) and~\ref{fig:failed_samples}(d), we illustrate two negative multimodal sample pairs that are accepted by the framework at false rejection rates (FRR) of $10^{-4}$. Similarly, we increase the FRR for these sample sets and these pairs are rejected at FRR = $5\times 10^{-3}$ and $10^{-2}$, respectively.

\begin{figure}[t]
\begin{center}
\includegraphics[width=.98\linewidth]{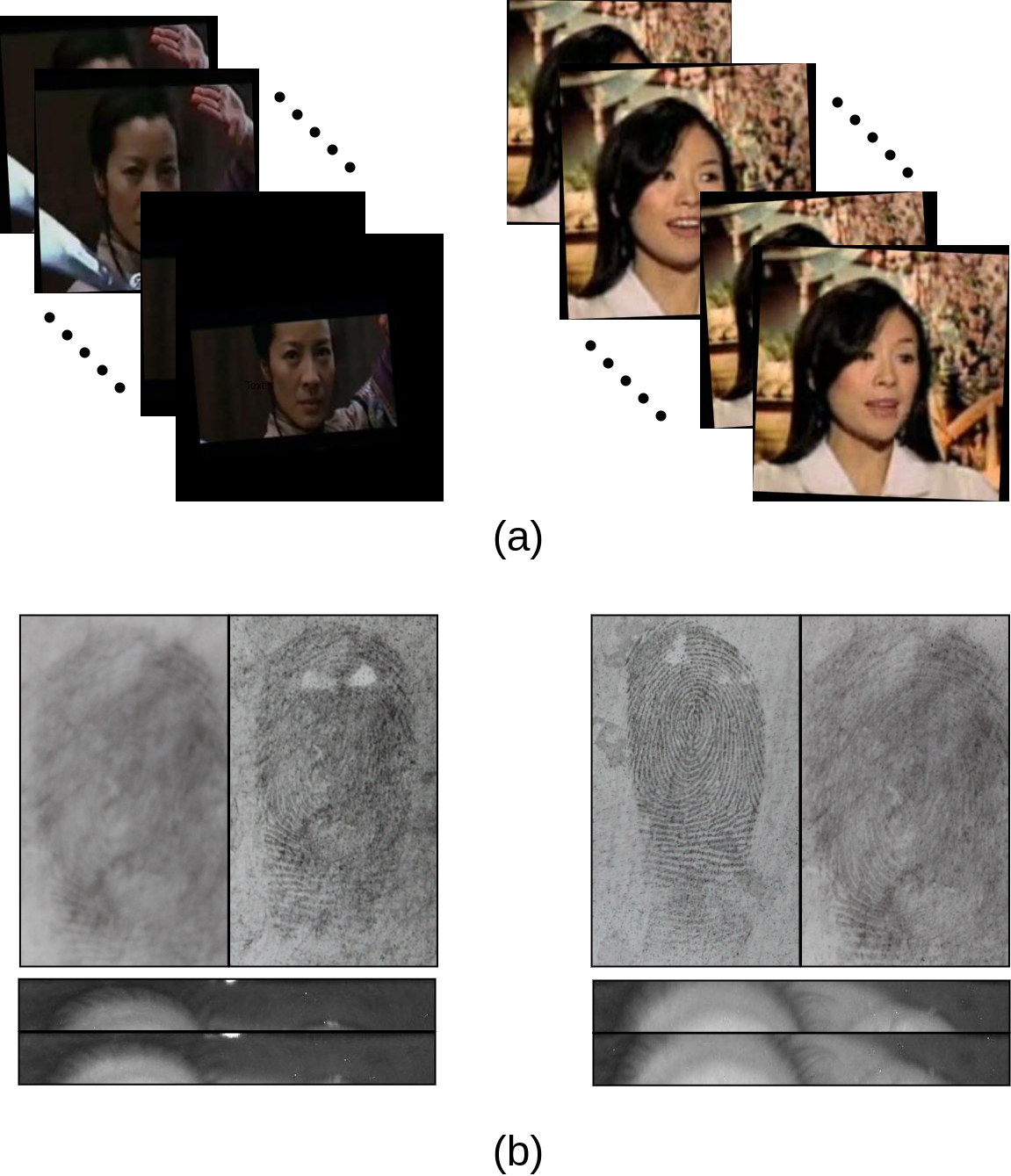}
\end{center}
\caption{Failed verification for the third multimodal dataset, (a) false rejection at FAR = $5\times 10^{-4}$ for the pairs of video frames and (b) correct verification at the same threshold when the iris and fingerprint samples are added to the multimodal sample sets.}
\label{fig:failed_samples_2}
\end{figure}

For the third multimodal dataset which consists of multiple samples of faces, irises captured from distance, and latent fingerprints, we study the performance when adding iris and fingerprint samples to a pair of incorrectly verified sample sets of face images. As presented in Fig.~\ref{fig:failed_samples_2}, the pairs of samples that we study include two sets of frames from the same class that are rejected by the framework at FAR = $5\times 10^{-4}$. However, by adding two latent fingerprint and two iris samples to each of the multimodal face sample sets, the framework successfully accepts the sample set at the same mentioned FAR. It might be noted that these fingerprint and iris samples are not independently verified correctly by the framework at the same threshold. 

One observation through our experiments is that, as discussed earlier in this section, although some of the fingerprint and iris samples are visually considered of very bad quality, the preprocessing plays a crucial role in their verification. We can also observe that the majority of the samples our framework fails to verify correctly are of very low quality or far from their class center. Another observation in these experiments is that the incorrectly accepted pairs of multimodal sample sets which pass the hardest FRR values are the pairs of very low quality. This is aligned with the common observation in image recognition that, in the representation domain, very low quality samples are relatively closer to the mass centers of the training set compared to the mass center of their own class~\cite{tokozume2018between,zhang2017mixup}.   

\section{Conclusions}
In this paper, we presented a quality-aware fusion with feature-level fusion for multi-sample multimodal recognition using modalities of face, iris, and fingerprint consisting of two blocks to handle different multi-sample multimodal scenarios. The first block provides high-level representation of each modality considering the quality of the samples in that modality. The second block provides a multimodal representation for the input multimodal sample set, while utilizing the inter-modality quality of modalities. The network is trained using the proposed multimodal separability loss, while the multimodal network compactness loss alleviates the over-fitting caused by the over-parametrization of multimodal networks. To study the performance of the proposed framework, as well as loss functions proposed in this paper, we consider three real-world multimodal and eight unimodal datasets. The proposed framework is trained end-to-end in a weakly-supervised fashion without any quality score supervision. The expectation of the inter-modality quality on each multimodal test dataset represents the importance of the modalities in the recognition task and is aligned with the unimodal performance of the framework. Compared to state-of-the-art algorithms, we demonstrated that the proposed architecture significantly improves the multimodal performance by estimating the quality of the samples. The proposed quality-aware fusion can be adapted to different preprocessing algorithms, feature extraction methods, and loss functions for recognition and several other multimodal applications such as access control security system, passport control, and unlocking the smartphones.

\begin{center}
ACKNOWLEDGEMENT
\end{center}
This material is based upon a work supported by the Center for Identification Technology Research and the National Science Foundation under Grant $\#1650474$.

\bibliographystyle{IEEEtran}
\bibliography{bib_TBIOM}

\begin{IEEEbiography}[{\includegraphics[width=1in,height=1.25in,clip,keepaspectratio]{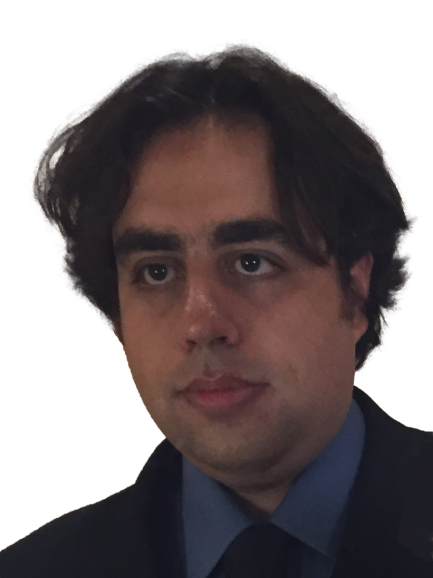}}]
{Sobhan Soleymani} received his B.Sc. degree in Electrical Engineering from University of Tehran and his M.Sc. degree in Electrical Engineering from \'Ecole Polytechnique F\'ed\'erale de Lausanne (EPFL). He is graduated with the Ph.D. degree in Electrical Engineering from the Lane Department of Computer Science and Electrical Engineering, West Virginia University in 2021. He has served as reviewer for IEEE Transactions on Biometrics, Behavior, and Identity Science (TBIOM), IEEE Transactions on Neural Networks and Learning Systems (TNNLS), Neural Information Processing Systems (NeurIPS), International Conference on Learning Representations (ICLR), International Joint Conference on Biometrics (IJCB), and Winter Conference on Applications of Computer Vision (WACV). His areas of interest are computer vision, machine learning, signal and image processing, and their application in biometrics. 
\end{IEEEbiography}

\begin{IEEEbiography}[{\includegraphics[width=1in,height=1.25in,clip,keepaspectratio]{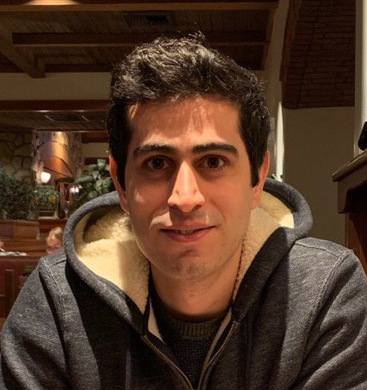}}]
{Ali Dabouei} received his master of science in electrical engineering from Sharif University of Technology, Tehran, Iran. He has been pursuing his doctoral studies since 2017 at West Virginia University, USA. His area of research includes machine learning, deep learning, and their applications in computer vision and biometrics. His research received best student paper awards in IEEE International Conference on Biometrics (BTAS), 2018, and IAPR International Conference on Biometrics (ICB) 2018. He has authored over 20 publications including journals and peer-reviewed conferences. He has also served as the reviewer for prestigious venues such as IEEE Transactions on Neural Networks and Learning Systems (TNNLS), International Conference on Computer Vision and Pattern Recognition (CVPR), and International Conference on Computer Vision (ICCV). 
\end{IEEEbiography}

\begin{IEEEbiography}[{\includegraphics[width=1in,height=1.25in,clip,keepaspectratio]{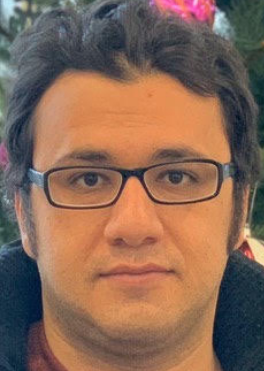}}]
{Fariborz Taherkhani} is currently a postdoctoral researcher at the Robotics Institute in Carnegie Mellon University, Pittsburgh, PA, USA. He received his PhD in computer science from West Virginia University, Morgantown, WV, USA.  He received his BSc. degree in computer engineering from National University of Iran, Tehran, Iran, and his MSc. degree in computer engineering from Sharif University of Technology, Tehran, Iran. His research interests include machine learning, computer vision, biometrics, and image retrieval.
\end{IEEEbiography}

\begin{IEEEbiography}[{\includegraphics[width=1in,height=1.25in,clip,keepaspectratio]{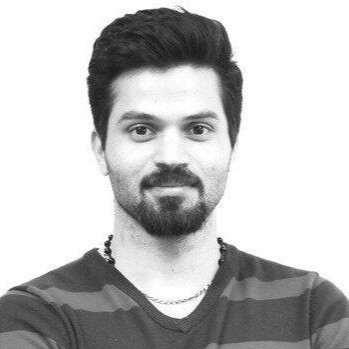}}]
{Seyed Mehdi Iranmanesh} received B.Sc. and M.Sc. degrees in Computer engineering and Artificial Intelligence from Iran University of Science and Technology, Tehran, Iran, in 2008 and 2011, respectively. He graduated in computer science from West Virginia University in 2021. His research interests include computer vision and machine learning. 
\end{IEEEbiography}

\begin{IEEEbiography}[{\includegraphics[width=1.25in,height=1.25in,clip,keepaspectratio]{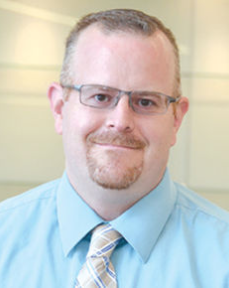}}]{Jeremy Dawson,} Associate Professor, joined the Lane Department of Computer Science and Electrical Engineering in the Fall of 2015. His background is in microelectronics and nanophotonics, and he has extensive experience in complex, multi-domain system integration and hardware system implementation. His current biometrics research efforts are focused on the creation of large-scale biometrics datasets that can be used to evaluate sensor operation and other human factors, as well as apply novel machine learning algorithms to solve human identification challenges.  Dr. Dawson also has extensive experience in micro and nanophotonic sensor platforms. His research in biosensors led to the identification of a need for new signal processing methodologies for DNA systems, which resulted in the first molecular biometrics (DNA) project funded through the WVU Center for Identification Technology Research (CITeR), an NSF IUCRC.
\end{IEEEbiography}

\begin{IEEEbiography}[{\includegraphics[width=1.25in,height=1.25in,clip,keepaspectratio]{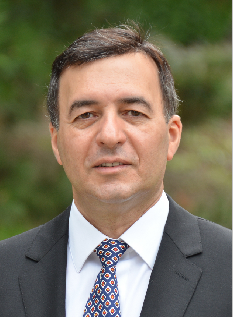}}]{Nasser M. Nasrabadi} (S’80 – M’84 – SM’92 – F’01) received the B.Sc. (Eng.) and Ph.D. degrees in electrical engineering from the Imperial College of Science and Technology, University of London, London, U.K., in 1980 and 1984, respectively. In 1984, he was with IBM, U.K., as a Senior Programmer. From 1985 to 1986, he was with the Philips Research Laboratory, New York, NY, USA, as a member of the Technical Staff. From 1986 to 1991, he was an Assistant Professor with the Department of Electrical Engineering, Worcester Polytechnic Institute, Worcester, MA, USA. From 1991 to 1996, he was an Associate Professor with the Department of Electrical and Computer Engineering, State University of New York at Buffalo, Buffalo, NY, USA. From 1996 to 2015, he was a Senior Research Scientist with the U.S. Army Research Laboratory. Since 2015, he has been a Professor with the Lane Department of Computer Science and Electrical Engineering. His current research interests are in image processing, computer vision, biometrics, statistical machine learning theory, sparsity, robotics, and neural networks applications to image processing. He is a fellow of ARL and SPIE and has served as an Associate Editor for the IEEE Transactions on Image Processing, the IEEE Transactions on Circuits, Systems and Video Technology, and the IEEE Transactions on Neural Networks.
\end{IEEEbiography}
\end{document}